\documentclass[sigconf]{acmart}

\usepackage{diagbox}
\usepackage{slashbox}
\usepackage{xspace}
\usepackage{wrapfig}
\usepackage[utf8]{inputenc} 
\usepackage[T1]{fontenc}    
\usepackage{hyperref}       
\hypersetup{hidelinks}
\usepackage{url}            
\usepackage{booktabs}       
\usepackage{amsmath}
\usepackage{amsthm}
\usepackage{amsfonts}       
\usepackage{nicefrac}       
\usepackage{microtype}      
\usepackage{xcolor}         
\usepackage{colortbl}
\usepackage{amsmath}
\usepackage{wrapfig}
\usepackage{graphicx}
\usepackage{makecell}
\usepackage{tabularx}
\usepackage{enumitem}
\usepackage{array} 
\usepackage{multirow}
\usepackage{caption}
\usepackage{subfigure}
\usepackage{bm}
\usepackage{tabularx}
\usepackage{framed}
\usepackage{xcolor}
\usepackage{stfloats} 
\usepackage{multirow}
\usepackage{xspace}
\usepackage[normalem]{ulem}
\useunder{\uline}{\ul}{}
\usepackage{natbib}

\AtBeginDocument{%
  }

\copyrightyear{2025}
\acmYear{2025}
\setcopyright{acmlicensed}\acmConference[SIGSPATIAL '25]{The 33rd ACM
International Conference on Advances in Geographic Information
Systems}{November 3--6, 2025}{Minneapolis, MN, USA}
\acmBooktitle{The 33rd ACM International Conference on Advances in
Geographic Information Systems (SIGSPATIAL '25), November 3--6, 2025,
Minneapolis, MN, USA}
\acmDOI{10.1145/3748636.3762727}
\acmISBN{979-8-4007-2086-4/2025/11}

\begin{document}

\title{Space-aware Socioeconomic Indicator Inference with Heterogeneous Graphs}

\author{Xingchen Zou}
\authornote{Both authors contributed equally to this research.}
\affiliation{%
  \institution{The Hong Kong University of Science and Technology (Guangzhou)}
  \city{Guangzhou}
  \country{China}
}
\email{xzou428@@connect.hkust-gz.edu.cn}

\author{Jiani Huang}
\authornotemark[1]
\affiliation{%
  \institution{The Hong Kong Polytechnic University}
  \city{Hong Kong}
  \country{China}}
\email{jianihuang01@gmail.com}

\author{Xixuan Hao}
\affiliation{%
  \institution{The Hong Kong University of Science and Technology (Guangzhou)}
  \city{Guangzhou}
  \country{China}
}
\email{xhao390@connect.hkust-gz.edu.cn}

\author{Yuhao Yang}
\affiliation{%
 \institution{University of Hong Kong}
 \city{Hong Kong}
 \country{China}}
\email{yuhao-yang@outlook.com}

\author{Haomin Wen}
\affiliation{
  \institution{Carnegie Mellon University}
  \city{Pittsburgh}
  \state{Pennsylvania}
  \country{USA}}
\email{wenhaomin.whm@gmail.com}

\author{Yibo Yan}
\affiliation{%
  \institution{The Hong Kong University of Science and Technology (Guangzhou)}
  \city{Guangzhou}
  \country{China}
}
\email{yanyibo70@gmail.com}

\author{Chao Huang}
\affiliation{
 \institution{University of Hong Kong}
 \city{Hong Kong}
 \country{China}}
\email{chaohuang75@gmail.com}

\author{Chao Chen}
\affiliation{%
  \institution{Chongqing University}
  \city{Chongqing}
  \country{China}}
\email{cschaochen@cqu.edu.cn}

\author{Yuxuan Liang}
\authornote{Y. Liang is the corresponding author. Email: yuxliang@outlook.com}
\affiliation{%
  \institution{The Hong Kong University of Science and Technology (Guangzhou)}
  \city{Guangzhou}
  \country{China}
}
\email{yuxliang@outlook.com}

\renewcommand{\shortauthors}{Zou et al.}
\newcommand{\model}{GeoHG\xspace}

\begin{abstract}
  Regional socioeconomic indicators are critical across various domains, yet their acquisition can be costly. Inferring global socioeconomic indicators from a limited number of regional samples is essential for enhancing management and sustainability in urban areas and human settlements. Current inference methods typically rely on spatial interpolation based on the assumption of spatial continuity, which does not adequately address the complex variations present within regional spaces. In this paper, we present GeoHG, the first space-aware socioeconomic indicator inference method that utilizes a heterogeneous graph-based structure to represent geospace for non-continuous inference. Extensive experiments demonstrate the effectiveness of GeoHG in comparison to existing methods, achieving an $R^2$ score exceeding 0.8 under extreme data scarcity with a masked ratio of 95\%. The code and data are available at \url{https://github.com/CityMind-Lab/GeoHG}.
\end{abstract}




\maketitle

\section{Introduction}

\begin{figure}[!t]
    \centering
    \includegraphics[width=0.95\linewidth]{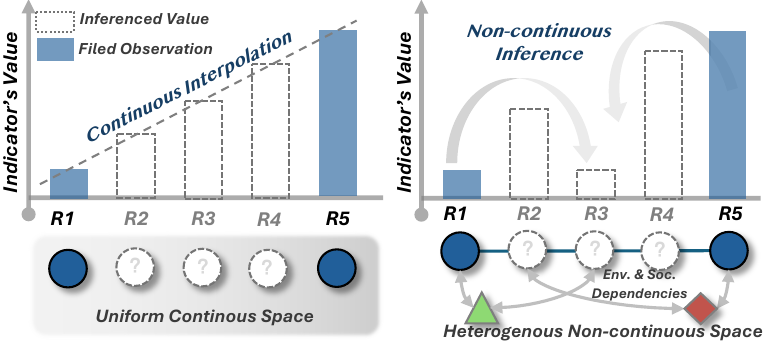}
    \vspace{-1.5em}
    \caption{Insight for space-aware socioeconomic indicator inference.}
    \label{fig:intro_geospace}
    \vspace{-1em}
\end{figure}

Our living space, as a complex system, reflects the interplay between natural laws and social dynamics. Socioeconomic indicators are designed to quantify the societal and economic dynamics of the space and contribute significantly to various domains such as the environment, management, sustainability, and urban computing~\citep{zou2024deep,xie2020urban,hao2025unlocking,zou2025fine}. In practice, these indicators are typically acquired through field investigations (e.g., population census) or deployed equipment (e.g., air monitoring station), with limited budgets resulting in a number of ground truth samples that is much lower than the global demand~\citep{costa2012systematic,ross2003survey}. In this context, \textbf{Socioeconomic Indicator Inference} aims to \emph{estimate the values of uninvestigated regions based on limited observations} in order to provide a more uniform and comprehensive quantification of the overall geospace.

In the past decade, most socioeconomic indicator inference methods, such as the kriging~\citep{brus2007optimization} method, have been developed using geometric approaches for spatial interpolation~\citep{mitas1999spatial}. While feasible, these methods depend on key assumptions of spatial continuity, spatial stationarity, and isotropic spatial autocorrelation, which imply that the events of the considered spatial indicators are distributed uniformly across the region, regardless of geographical directions or barriers~\citep{mitas1999spatial,tan2014comparative}. However, the actual distribution of these indicators is non-continuous and influenced by multiple dependencies. For instance, two adjacent regions may have different population densities due to variations in land environment~\citep{ross2003survey}, while regions along the same river basin may exhibit similar GDP patterns despite being separated by distance~\citep{li2024gdp,wang2019discovery}. Consequently, the interpolated values can significantly diverge from the real scenario. Additionally, the available samples are typically non-uniformly distributed in space, resulting in poor or even ineffective performance of traditional interpolation methods in sparse areas~\citep{li2011review}.

\begin{figure*}
    \centering
    \includegraphics[width=0.95\linewidth]{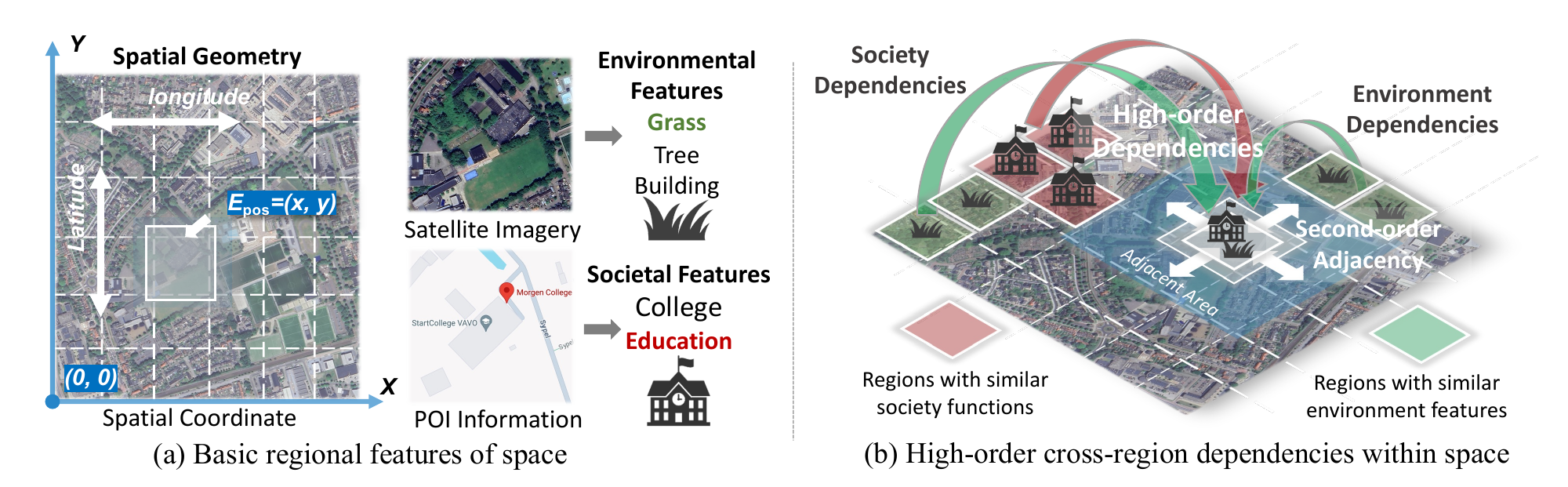}
    \vspace{-1.5em}
    \caption{Illustration of geospatial features and dependencies for spare-aware indicator inference.}
    \vspace{-1em}
    \label{fig:intro}
\end{figure*}

To address this issue, it is important to integrate artificial intelligence to uniformly represent features and non-Euclidean relationships within geospace using geospatial data. This representation can then enhance global indicator inference in non-continuous and non-uniform scenarios (see Fig. \ref{fig:intro_geospace}). While significant research on region representation exists~\citep{liu2023urbankg,wang2020urban2vec,hao2025nature,zhang2024deepmeshcity,mai2024opportunities}, most studies assume redundant samples and train models on large datasets to predict individual samples. The design of a method that comprehensively represents geospatial data for indicator inference remains unexplored.

Introducing such methods presents two main challenges. \textit{First, the data reliance for geospace representation must be robust}. Existing region representation methods may utilize various data sources, such as trajectories, street-view imagery, or road networks, which might not always be available. Although some studies leverage globally available satellite imagery to generate general spatial representations, their specified vision encoders may perform poorly when trained with only a few samples. 
\textit{Second, comprehensively representing environmental and societal dependencies in geospatial data is non-trivial}. As shown in Figure \ref{fig:intro}, space-aware inference should jointly consider spatial geometry, environmental, and societal dependencies, which extend beyond basic regional features and second-order adjacency to include high-order dependencies across regions. For instance, a certain region may be influenced jointly by a group of non-adjacent regions due to shared geographic or socioeconomic functions.

To address these challenges, we present GeoHG, the first space-aware socioeconomic indicator inference method that utilizes a heterogeneous graph-based structure to jointly represent geospace from spatial geometry, environmental, and societal perspectives for non-continuous inference. For robustness, we employ only satellite imagery and points of interest (POI) to represent the environmental and societal features of the space, as these are among the most accessible data sources. Furthermore, to overcome the limitations of traditional vision encoders, we directly perform semantic segmentation on the satellite imagery to distinguish geo-entities such as water bodies, vegetation, and buildings, enabling a more general and concise representation. For comprehensive geospace representation, we introduce a novel heterogeneous graph that integrates spatial geometrical relations with encoded environmental and societal information. This graph structure effectively captures the non-Euclidean local relationships within the space. Additionally, our heterogeneous formulation between region-entity associations accounts for the high-dimensional societal and environmental dependencies across regions. By uniformly representing the space, we can conduct inference on the graph for the entire region with consistent accuracy, regardless of the distribution of available samples. 
Overall, our contribution can be summarized as follows:

\begin{itemize}[leftmargin=*]

\item \textit{Space-aware Inference.} We introduce a new space-aware method for socioeconomic indicator inference that incorporates a joint representation of spatial geometry, environmental, and societal relations. To the best of our knowledge, this is the first work that integrates heterogeneous graph-based space representation for socioeconomic indicator inference.
 
\item \textit{Heterogeneous Graph for Joint Space Representation.} We develop a heterogeneous graph structure to model geospace using widely accessible data sources, capturing high-order societal and environmental dependencies through heterogeneous connections.

\item \textit{Empirical Evidence.} Our experiments demonstrate that GeoHG outperforms existing interpolation methods and region representation models across various downstream tasks. It achieves an $R^2$ score exceeding 0.8 under extreme data scarcity with a masked ratio of 95\% and performs effectively on inference with highly unevenly distributed samples.

\end{itemize}

\begin{figure*}
    \centering
    \includegraphics[width=0.95\linewidth]{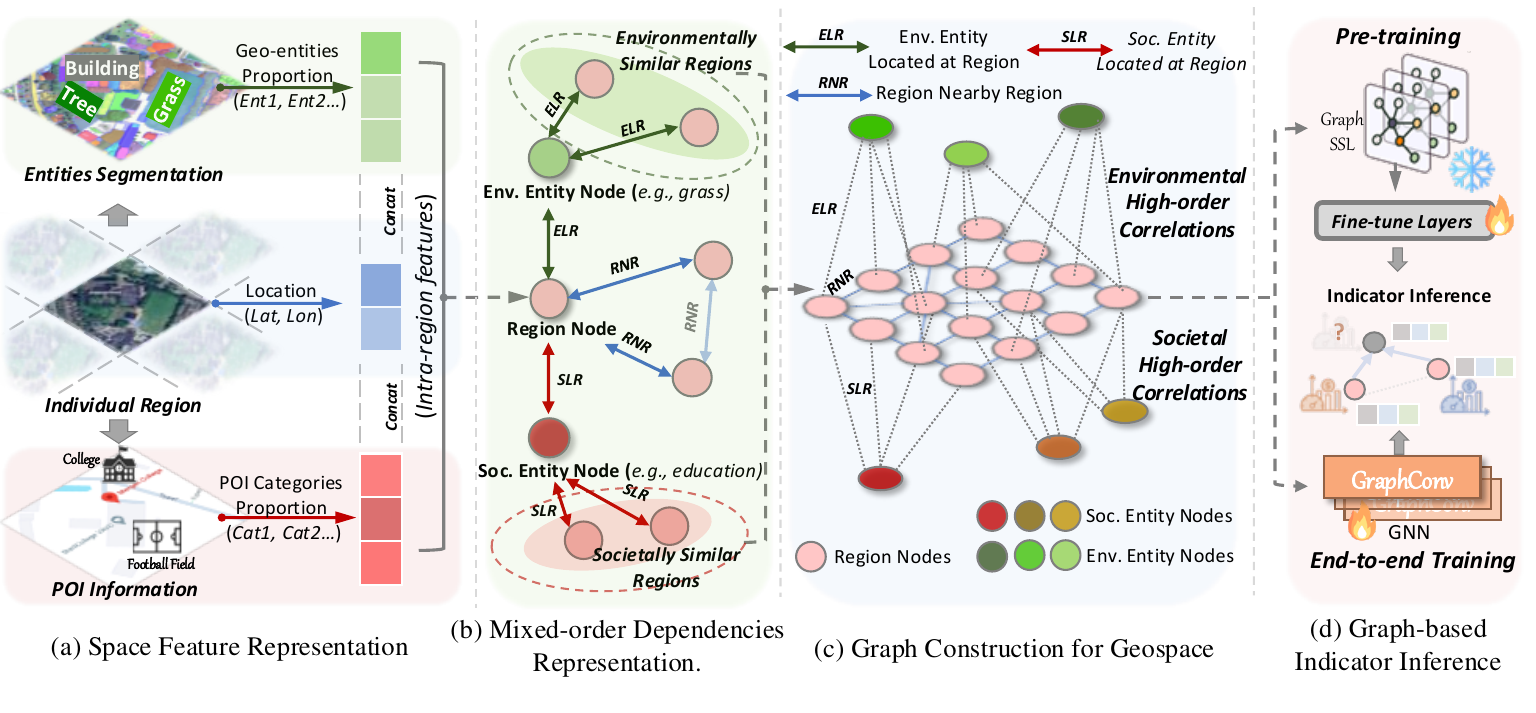}
    \vspace{-2em}
    \caption{Overall framework of the proposed GeoHG.}
    \label{fig:framework}
\end{figure*}

\section{Formulation}

An investigated geospace $\mathcal{R}$ typically comprises multiple regions denoted as $\mathcal{R}=\left\{r_1, r_2, \cdots, r_N\right\}$. These regions exhibit unique environmental and societal characteristics, often reflected through various data sources within them. In our study, we represent $r_i \in \mathcal{R}$ through the following aspects:
\begin{itemize}[leftmargin=*]
    \item \textit{\textbf{Spatial Geometry}} represents the spatial location and geometrical relations of $r_i$. The spatial location can be represented by its center coordinate $(lon^i, lat^i)$, and the geometrical relations can be captured by computing the distance between each pair of regions $r_i$.

    \item \textit{\textbf{Satellite Imagery}} captured through ongoing earth observation by satellites, provides an overview of the environmental features of $\mathcal{R}$. Each satellite image of a region $r_i$ can be expressed as $S_i \in \mathbb{R}^{H \times W \times L}$, where \( H \times W \) denotes the image's size and \(L\) represents the number of color channels. 

    \item \textit{\textbf{Point of Interest}} refers to a specific location or geographic point of significance to society, which includes campuses, businesses, or any other notable locations. POI can be represented as $P = (x, y, c)$, where \( x \) and \( y \) are the geographical coordinates, and \( c \) is the societal category.
\end{itemize}

\noindent \textbf{Problem Statement.} Given a geospace $\mathcal{R}$ with corresponding satellite images \(\mathbf{S} =\left\{S_1, S_2, \cdots, S_N\right\}\) and a set of POIs \(\mathbf{P}_i=\left\{P_1, P_2, \cdots, P_M\right\}\), our goal is to construct a graph \( \mathcal{G}\) that captures the spatial and contextual characteristics of the space. Within the space graph \( \mathcal{G} \), our approach aims to generate precise estimates of socioeconomic indicators $\mathcal{Y}$ for the unlabeled regions $\mathcal{R}_{mask}\subseteq\mathcal{R}$ from the available regions $C_{\mathcal{R}_{mask}} \mathcal{R}$. This process can be formulated as:
\begin{equation}\nonumber
\left(\hat{\mathcal{Y}}^{r_1}, \hat{\mathcal{Y}}^{r_2}, \cdots, \hat{\mathcal{Y}}^{r_k}\right) \leftarrow \mathcal{F}_\theta\left(\mathcal{G}|\mathcal{Y}^{r_{k+1}}, \mathcal{Y}^{r_{k+2}}, \cdots, \mathcal{Y}^{r_N}\right),
\end{equation}
where $\{r_1,r_2,\cdots,r_k\}\in\mathcal{R}_{mask}$ and we use $\mathcal{M} = k/N$ as the masked ratio. $\mathcal{F}_\theta$ is the inference model.

\section{Methodology}

\label{Sec:method}

As illustrated in Figure~\ref{fig:framework}, our GeoHG framework comprises four major components:

\begin{itemize}[leftmargin=*]
    \setlength{\itemsep}{3pt}
    \setlength{\parsep}{0pt}
\setlength{\parskip}{0pt}
    \item \textbf{Space Feature Representation.} We first extract basic space features from satellite imagery and POI data, considering the proportion of geo-entities, POI category distribution, and the regions' positions. The proportions of geo-entities are derived via semantic segmentation of satellite imagery and clustering of POI by categories.
    \item \textbf{Mixed-order Dependencies Representation.} To model cross-region spatial and socioeconomic dependencies, we construct a heterogeneous graph connection that reflects the mixed-order relationships from spatial, environmental and societal perspectives. 
    \item \textbf{Graph Construction for Geospace.} We combine the basic features and cross-region dependencies using a comprehensive graph. The basic features are incorporated into the node vectors, while the mixed-order relations are expressed through the edges and their corresponding weights.
    \item \textbf{Graph-based Indicator Inference.} We introduce two approaches to inference regional indicators on the graph: self-supervised graph learning for creating region embeddings for light-weight deployment and end-to-end training to directly inference on the graph for peak performance.

\end{itemize}

\subsection{Space Feature Representation}
\label{sec:intra}

To effectively encode spatial, environmental, and societal signals into feature embeddings, we extract multi-modal geo-contextual data including satellite imagery, POI data, and the location of the regions.

\noindent \textbf{Spatial Geometric Embedding. }In line with industrial conventions, the finest spatial grain of socioeconomic indicators is limited to a $1 \text{ km}^2$ scale~\citep{yan2023urban,hao2024urbanvlp}. We divide the overall geospace into multiple $1 \text{ km} \times 1 \text{ km}$ spatial grids, using a coordinate system with the origin at $(lon^0, lat^0)$. The abscissa $x_R$ and ordinate $y_R$ of the region $R$ are then utilized as spatial information $E_{pos}(x,y)$. In particular, for any region $R_i$ in the target space, we have:
\begin{equation}
    E^{i}_{pos}=(lon^i-lon^0, lat^i-lat^0)\Delta*D^{-1}
\end{equation}
where $\Delta$ is the transform vector decided by the coordinate system. $D$ denotes the scale of the grids, and for $1\textrm{km} \times 1\textrm{km}$ grids, $D=1$. It is noteworthy that the $1 \text{ km}^2$ scale can be easily downsampled to other scales for improved compatibility.

\noindent \textbf{Environmental Feature Embedding}:
Conventional vision encoders are designed and trained on natural images, which are considerably different from satellite images. Understanding the complex and highly specialized geo-semantic information in satellite imagery is a challenging task. However, satellite images display a significant level of structured organization concerning semantic content, in contrast to natural images, which contain a vast array of diverse information. To accurately analyze a satellite image, one needs to focus solely on the geological entities it includes, the extent of these entities, and their positions within the geospace. Therefore, using an entity segmentation method to encode satellite images is promising, as it can directly offer crucial information regarding the entities and their spatial distribution.

\begin{figure*}[t!]
    \centering
    \includegraphics[width=0.8\linewidth]{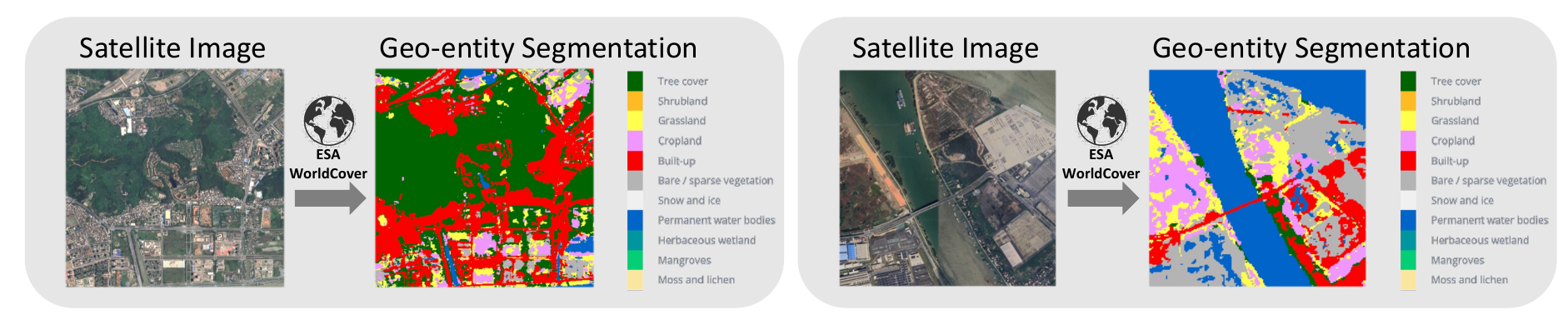}
    \vspace{-1.5em}
    \caption{Geo-entity Segmentation for Satellite Images based on ESA WorldCover.}
    \label{fig:seg1}
\end{figure*}

Inspired by this, we unitize the European Space Agency (ESA) with ESA WorldCover Dataset~\citep{zanaga2022esa}, elaborate in Figure \label{fig:seg1}, and design a segmentation-based process to encode the satellite images instead of the most commonly used visual encoders. We utilize satellite imagery to mine the environmental features and expect the satellite imagery encoding to be efficient and concise. Given a satellite image $S_i$ of a region of interest $R_i$, we conduct semantic segmentation
to obtain a series of geo-entities $\{entity_{1},entity_{2}, ... ,entity_{j}\}_{j=1}^J $ within that region, such as developed areas, grass, and trees based on ESA Worldcover, leveraging their area proportion as environmental feature embedding as below:
\begin{equation}
   E_{env}=\{p_{ent_1},p_{ent_2}, ... ,p_{ent_j}\}_{j=1}^J 
\end{equation}
where $p_{ent_j} = \frac{A_j}{A_R}$, and $A_j$ is the area occupied by $entity_j$, $A_R$ is the area of region $R_i$, $J$ is the number of environmental entity types. 

\noindent \textbf{Societal Feature Embedding}:
To effectively represent the societal feature of a region, given POI dataset $D_{POI}$ and the region of interest $R_i$, we firstly find all points $D^{R_i}_{POI}$($poi_1, poi_2,..., poi_j)$ which are located at $R_i$. Then, we count the proportion of different $K$ POI categories in the region $R_i$ to get $E_{soc}^{\prime}$ as below:
\begin{equation}\nonumber
  E_{soc}^{\prime} = \{p_{poi_1},p_{poi_2}, ... , p_{poi_k}\}_{k=1}^K, \quad p_{\text{poi}_k} = \frac{|C_k \cap D_{POI}^{R_i}|}{|D_{POI}^{R_i}|}
\end{equation}
where $K$ is the number of POI categories, $C_k = \{poi_j \in D_{POI} \mid \text{category}(poi_j) = k\}$. However, directly employing the proportion of different POI categories $E_{soc}^{\prime}$ as societal features can still be problematic since it cannot distinguish between areas with high POI density (usually with a higher degree of socialization) and those with relatively low POI density. Typically, a higher POI density region is more significantly influenced by its social characteristics. Thus, we convert $E_{soc}^{\prime}$ into the final societal feature embedding $E_{soc}$ by applying a \textit{Social impact factor} $f$ as shown below:
\begin{equation}
\label{social_factor}
     E_{soc} = f\cdot E_{soc}^{\prime}, \quad f = log( D^{R_i}_{POI} +1)
\end{equation}




\subsection{Mixed-order Dependencies Representation.}
\label{inter}

Region indicators are also significantly influenced by various cross-region dependencies. Our objective is to jointly model pairwise second-order spatial relationships between regions, as well as higher-order dependencies among groups of regions that exhibit similar environmental or societal characteristics, using a heterogeneous graph structure.

\begin{figure}[!h]

    \centering
    \includegraphics[width=0.9\linewidth]{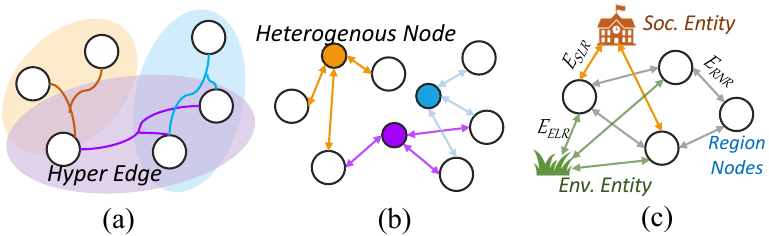}
    \vspace{-1em}
    \caption{Illustration of high-order relations representation in GeoHG. Given a hypergraph (a) with high-order relations, we reconstruct it into a heterogeneous graph (b) by adding nodes as higher-order connection channels and get the final representation (c) in $\mathcal{G}$.}
    \vspace{-1em}
    \label{fig:hyper}
\end{figure}

According to hypergraph theories \citep{liang2022mixed,feng2019hypergraph}, we employ a heterogeneous graph structure to represent cross-region relations. As shown in Figure \ref{fig:hyper}, we utilize heterogeneous nodes in the graph structure as transfer nodes to facilitate high-order message passing (equivalent to hypergraph edges). Given a geospace $\mathcal{R}$, we construct a weighted heterogeneous graph $\mathcal{G}$, which contains three types of nodes: \textit{region nodes}, \textit{environmental entity nodes}, and \textit{societal entity nodes}. Specifically, \textit{region nodes} correspond to the grid cells of $\{r_1, r_2, \cdots, r_N\}$; \textit{environmental entity nodes} represent the $J$ distinct geo-entity classes detected from satellite image $S_i$; and \textit{societal entity nodes} comprise the $K$ different POI categories. The model structure can be found in Figure \ref{fig:framework}. We will elaborate on how second-order and high-order information is represented.

\noindent \textbf{Second-Order Relation Representation}: Second-order relations capture the pairwise spatial adjacency between regional nodes in $\mathcal{G}$. We construct undirected edges $\mathcal{E}_{RNR}$, \textit{Region Nearby Region (RNR)} between regional nodes whose corresponding grid cells are spatially adjacent in a $3\times3$ grid, thereby encoding these local second-order dependencies.
\begin{equation}
    \mathcal{E}_{RNR, m n}=1, \text{if } \mathcal{V}_{m} \text { and } \mathcal{V}_{n} \text{ are \textit{adjacent} in geospace}
\end{equation}

\noindent \textbf{High-Order Relation Representation}: Multi-view graphs \citep{geng2019spatiotemporal,zou2024deep} represent the relations between regions through additional edges and are confined to second-order regional connections. To fully capture the higher-order relations among urban regions, we explicitly model these higher-order interactions utilizing heterogeneous environmental entity nodes $\mathcal{V}_{env}$ and societal entity nodes $\mathcal{V}_{soc}$ within $\mathcal{G}$. Specifically, we construct heterogeneous weighted edges $\mathcal{E}_{ELR}$ and $\mathcal{E}_{SLR}$ through $\mathcal{V}_{env}$ and $\mathcal{V}_{soc}$ to associate each regional node with its constituting environmental entities (e.g., water, vegetation) and societal entities (e.g., educational, commercial POIs). For $p_{ent_{it}} \geq \theta_{Env}$ or $f \cdot p_{poi_{ik}} \geq \theta_{Soc}$:
\begin{equation}
    \mathcal{E}_{ELR,i j}=1\cdot p_{ent_{ij}},\quad
    \mathcal{E}_{SLR,i k}=1 \cdot f \cdot p_{poi_{ik}}
\end{equation}
where $p_{ent_{ij}}$ represents the area proportion of geo-entity $j$ at region $i$, $p_{poi_{ij}}$ denotes the quantity proportion of POI category $k$ at region $i$ and $f$ is the \textit{social impact factor} elaborated in Equation \ref{social_factor}. $\theta_{Env}$ and $\theta_{Soc}$ serve as hyperparameters designed to optimize the graph structure for conciseness and efficiency.
This edge formulation allows encoding higher-order relationships between regions exhibiting similar environmental/societal attributes, even if they lack spatial proximity. By combining the second-order spatial adjacency edges with these higher-order hyperedge associations, $\mathcal{G}$ holistically represents the mixed-order relational patterns within the region.



\subsection{Graph Construction for Geospace}
\label{integration}

After deriving the basic space features and the inter-regional dependencies encoded in the heterogeneous graph $\mathcal{G}$, we use a model-agnostic graph framework to integrate these information for a comprehensive space representation $E_{space}$. Specifically, this phase can be formulated as:
\begin{equation}
    E_{space} = \text{GNN}(\mathcal{G}[E_{pos},E_{env},E_{soc}])
\end{equation}
In the graph, each regional node $v$ in $\mathcal{G}$ is represented with an initial node feature corresponding to its regional feature $E_{pos},E_{env},E_{soc}$ elaborated in Sec \ref{sec:intra}. Note that for simplicity, instead of utilizing hypergraph neural networks~\citep{feng2019hypergraph, yang2022multi,xia2022self}, we model hyperedges with the form of complete subgraphs in $\mathcal{G}$.
Therefore, we apply Heterogeneous Graph Neural Networks (HGNNs) over $\mathcal{G}$ to update the node representations by iteratively aggregating and transforming multi-hop neighborhood information:
\begin{align}
    \mathbf{x}_v^{(l+1)} &= \text{HGNN}^{(l)}\left(\mathbf{x}_v^{(l)}, \mathcal{N}(v), \mathcal{G}_i\right), \mathbf{x}_v^{(l)} \in \mathbb{R}^d
\end{align}
where $\mathcal{N}(v)$ denotes the neighbors (spatial and higher-order) of node $v$ in $\mathcal{G}$. 
After $L$ layers of updates, the final node embedding $\mathbf{x}_v^{(L)}$ integrates cues from the basic features as well as mixed-order inter-region dependencies.


\subsection{Graph-based Indicator Inference}
\label{deployment}


To effectively address various practical demands, we present two inference strategies: \textit{self-supervised graph learning}, which learns fine-grained, generalizable region representations within the space, enabling lightweight deployment and transferability to diverse downstream tasks; and \textit{end-to-end training}, which allows for direct inference on the graph to achieve optimal performance.

\noindent \textbf{Self-Supervised Graph Learning}: Inspired by CLIP~\citep{radford2021learning}, we pre-train our model using a contrastive learning paradigm where we maximize the similarity between a region $\mathbf{R}_i$ and its corresponding positive region groups $\mathbf{C}_i$, while minimizing its similarity with other region groups in the batch. Specifically, we sample regional nodes adjacent to $\mathbf{R}_i$ as well as nodes exhibiting similar intra-regional features to construct $\mathbf{C}_i$. During training, we leverage a GNN model $\mathcal{M}_{pretrain}$ to obtain embeddings for $\mathbf{R}_i$ and all nodes in $\mathbf{C}_i$. The embeddings of the sampled positive regional nodes are further pooled to obtain the unified positive region representation $\mathbf{e}_j$. The process of generating embeddings for $\mathbf{R}_i$ and $\mathbf{C}_i$ is conducted as follows:
\begin{equation}
    \mathbf{e}_i = \mathcal{M}_{pretrain}(\mathbf{X}_i),\quad
    \mathbf{e}_j = \operatorname{Pool}\left(\mathcal{M}_{pretrain}(\mathbf{X}_j), {j \in \mathcal{C}_i} \right)
\end{equation}

We further define a similarity function $f_{\mathbf{score}}$ to measure the similarity between the representations $\mathbf{e}_i, \mathbf{e}_j$. $f_{\mathbf{score}}$ can be a simple dot product or a more complex metric and then employ an InfoNCE-based loss~\citep{oord2018representation} to conduct contrastive learning:
\begin{equation}
    \mathcal{L}_{pretrain}=\mathbb{E}[-\log \frac{\exp \left(f_{\mathbf{score}}\left(\mathbf{e}_i, \mathbf{e}_j\right)\right)}{\sum_{\forall i, n} \exp \left(f_{\mathbf{score}}\left(\mathbf{e}_i, \mathbf{e}_n\right)\right)}]
\end{equation}
where $f_{\mathbf{score}}(\mathbf{e}_i, \mathbf{e}_j)$ represents the score of positive pairs while $f_{\mathbf{score}}\left(\mathbf{e}_i, \mathbf{e}_n\right)$ refers to the scores of negative pairs. 

After obtaining the pre-trained model $\mathcal{M}_{pretrain}$ with pre-trained region embedding $\mathbf{E}_{pretrain}$ through self-supervised learning, we fine-tune it with a available samples by adding three trainable linear layers. The weights of these additional linear layers are updated during fine-tuning while the weights of $\mathcal{M}_{pretrain}$ and $\mathbf{E}_{pretrain}$ remain fixed:
\begin{equation}
 \hat{\mathbf{y}} = \text{MLP}(\mathbf{E}_{pretrain}) 
\end{equation}

\noindent \textbf{End-to-End Training}: For direct inference of socioeconomic indicators on the graph, the indicator inference can be regarded as a node label prediction task. In practice, we optimize all parameters from scratch using the task's supervised signal. We adopt the HGNN model $\mathcal{M}$ to obtain embeddings for a given region $i$, which are then fed into a three-linear layer regression head denoted as $\mathbf{Regressor}$. All parameters are updated by minimizing the Mean Square Error (MSE) loss on the available sample nodes. The overall process can be formulated as follows:
\begin{equation}\nonumber
    \hat{y}_i = \mathbf{Regressor}(\mathcal{M}(\mathbf{X}_i)), \quad
    \mathcal{L}_{\text{MSE}} = \frac{1}{N} \sum_{i=1}^{N} (\hat{y}_i - y_i)^2
\end{equation}


\section{Experiments}
\label{sec:experiment}


    

\subsection{Experimental Setup}

\subsubsection{Datasets.} The datasets used in this paper include satellite imagery, POI information and five region indicators (Population, GDP, Night Light, Carbon, $PM_{2.5}$) located in four representative cities in China: Beijing, Shanghai, Guangzhou, and Shenzhen. All indicators are provided at 1 $km^2$ scale by data servers.We employ five representative tasks located in four cities in China: Beijing, Shanghai, Guangzhou and Shenzhen. Population, GDP, and Night Light tasks reflect anthropogenic activities, whereas Carbon and Temperature tasks characterize the natural environment. We conduct data preprocessing according to~\citep{yan2023urban}. The detailed information of datasets is listed below:
\begin{itemize}[leftmargin=*]
    \item \textbf{Carbon Emissions}: This dataset incorporates anthropogenic CO2 emission estimates sourced from the Open Data Inventory (ODIAC) for the year 2022, spatially aligned with our 1 km$^2$ satellite image grids (emissions quantified in tons).
    \item \textbf{Population}: This dataset is obtained through WorldPop population distribution data for 2020, with counts representing the number of citizens per region.
    \item \textbf{GDP}: This dataset includes Gross Domestic Product (GDP) statistics reflecting China's regional economic development patterns from \citet{zhao2017forecasting}.
    \item \textbf{Night Light}: As a proxy for human activity intensity, a key driver of urban evolution, we leverage nighttime light imagery data from~\citet{zhong2022development} in 2020.
    \item $\bm{PM}_{2.5}$: The $PM_{2.5}$ dataset is sourced from ChinaHighPM2.5 dataset~\citep{pm25dataset}. This dataset combines ground-based observations, atmospheric reanalysis, emission inventory, and other techniques to obtain nationwide seamless ground PM2.5 data from 2000 to the present. The main scope is the entire China area, the spatial resolution is 1 km, the time resolution is daily, monthly, and yearly, and the unit is µg/m3.
    
\end{itemize}

\begin{table}[!b]
\centering

\caption{Dataset statistics.}
\vspace{-1em}
\label{tab:dataset_statistics}
\resizebox{1.0\columnwidth}{!}{
\begin{tabular}{c|c|c|c|c} 
\toprule
\multirow{2}{*}{\textbf{Dataset}} & \multicolumn{2}{c|}{\textbf{Coverage in Geospace}} & \multirow{2}{*}{\textbf{\makecell{Satellite \\ Image}}}  & \multirow{2}{*}{\textbf{\makecell{POI \\ Data}}}  \\

\cline{2-3}
& \textit{Bottom-left} & \textit{Top-right}
 \\ 
\hline
Beijing & 39.75°N, 116.03°E & 40.15°N, 116.79°E  & 4,277&709,232 \\
Shanghai & 30.98°N, 121.10°E & 31.51°N, 121.80°E & 5,292&808,957\\
Guangzhou & 22.94°N, 113.10°E & 23.40°N, 113.68°E & 8,540& 805,997 \\
Shenzhen & 22.45°N, 113.75°E & 22.84°N, 114.62°E &5,150&717,461 \\
\toprule
\end{tabular}
}
\end{table}


The overall dataset statistics can be listed in Table~\ref{tab:dataset_statistics}. We randomly \underline{mask 75\%} data as unlabeled samples for test, the available samples are divided to 80\% for train and 20\% for validation.

\begin{table*}
\captionsetup{font=small}
\caption{Socioeconomic indicators inference results ($\text{masked ratio} = 75\%$). The bold/underlined font means the best/the second-best result.}
\label{tab:results_overall}
\vspace{-2mm}
\setlength{\tabcolsep}{0.9mm}{}
\begin{small}
\scalebox{0.8}{
\begin{tabular}{c|c|ccc|ccc|ccc|ccc|ccc|ccc|ccc|ccc}
\toprule

\multicolumn{2}{c|}{Methods} & \multicolumn{3}{c|}{\model} &\multicolumn{3}{c|}{\model-SSL} & \multicolumn{3}{c|}{UrbanVLP} & 
\multicolumn{3}{c|}{GeoStructural} & 
\multicolumn{3}{c|}{PG-SimCLR} & 
\multicolumn{3}{c|}{UrbanCLIP} &
\multicolumn{3}{c|}{IDW} &
\multicolumn{3}{c}{UK} \\

\midrule

\multicolumn{2}{c|}{Metric} & \cellcolor{blue!3}$\text{R}^2$ &  \cellcolor{green!3}MAE &  \cellcolor{orange!2}RMSE & \cellcolor{blue!3}$\text{R}^2$ &  \cellcolor{green!3}MAE & \cellcolor{orange!2}RMSE & \cellcolor{blue!3}$\text{R}^2$ &  \cellcolor{green!3}MAE &  \cellcolor{orange!2}RMSE & \cellcolor{blue!3}$\text{R}^2$ &  \cellcolor{green!3}MAE &  \cellcolor{orange!2}RMSE & \cellcolor{blue!3}$\text{R}^2$ &  \cellcolor{green!3}MAE&   \cellcolor{orange!2}RMSE & \cellcolor{blue!3}$\text{R}^2$ &  \cellcolor{green!3}MAE &  \cellcolor{orange!2}RMSE &
\cellcolor{blue!3}$\text{R}^2$ &  \cellcolor{green!3}MAE & \cellcolor{orange!2}RMSE
&
\cellcolor{blue!3}$\text{R}^2$ &  \cellcolor{green!3}MAE &  \cellcolor{orange!2}RMSE
\\

\midrule

\multirow{6}{*}{\rotatebox{90}{Beijing}}
& Carbon & \cellcolor{blue!3}\textbf{0.894}    & \cellcolor{green!3}\textbf{0.112}  & \cellcolor{orange!2}\textbf{0.211} & \cellcolor{blue!3}\underline{0.887} &     \cellcolor{green!3}\underline{0.161} & \cellcolor{orange!2}\underline{0.224} & \cellcolor{blue!3}0.717 &     \cellcolor{green!3}0.363 & \cellcolor{orange!2}0.467 & \cellcolor{blue!3}0.705 &     \cellcolor{green!3}0.398 & \cellcolor{orange!2}0.482 & \cellcolor{blue!3}0.402  &     \cellcolor{green!3}0.651 & \cellcolor{orange!2}0.764 & \cellcolor{blue!3}0.604 &     \cellcolor{green!3}0.558 & \cellcolor{orange!2}0.598 & \cellcolor{blue!3}0.734   &     \cellcolor{green!3}0.367 & \cellcolor{orange!2}0.489 & \cellcolor{blue!3}0.745   &     \cellcolor{green!3}0.355 & \cellcolor{orange!2}0.454
\\

& Population & \cellcolor{blue!3}\textbf{0.834}    & \cellcolor{green!3}\textbf{0.271} & \cellcolor{orange!2}\textbf{0.351} & \cellcolor{blue!3}\underline{0.830} &     \cellcolor{green!3}\underline{0.282} & \cellcolor{orange!2}\underline{0.374} & \cellcolor{blue!3}0.695  &     \cellcolor{green!3}0.404 & \cellcolor{orange!2}0.513 & \cellcolor{blue!3}0.701  &     \cellcolor{green!3}0.402 & \cellcolor{orange!2}0.504 & \cellcolor{blue!3}0.451  &     \cellcolor{green!3}0.964 & \cellcolor{orange!2}1.117 & \cellcolor{blue!3}0.431  &     \cellcolor{green!3}0.552 & \cellcolor{orange!2}0.669 & \cellcolor{blue!3}0.689  &     \cellcolor{green!3}0.385 & \cellcolor{orange!2}0.517 & \cellcolor{blue!3}0.701   &     \cellcolor{green!3}0.381 & \cellcolor{orange!2}0.499
\\

& GDP & \cellcolor{blue!3}\textbf{0.617}    & \cellcolor{green!3}\textbf{0.336} & \cellcolor{orange!2}\textbf{0.567} & \cellcolor{blue!3}\underline{0.614}  &     \cellcolor{green!3}\underline{0.331} & \cellcolor{orange!2}\underline{0.581} & \cellcolor{blue!3}0.556  &     \cellcolor{green!3}0.416 & \cellcolor{orange!2}0.650 & \cellcolor{blue!3}0.567  &     \cellcolor{green!3}0.401 & \cellcolor{orange!2}0.609 &  \cellcolor{blue!3}0.247 &     \cellcolor{green!3}0.768 & \cellcolor{orange!2}1.254 & 
\cellcolor{blue!3}0.315  &     \cellcolor{green!3}0.539 & \cellcolor{orange!2}0.864 &
\cellcolor{blue!3}0.555  &     \cellcolor{green!3}0.414 & \cellcolor{orange!2}0.619 & \cellcolor{blue!3}0.411   &     \cellcolor{green!3}0.621 & \cellcolor{orange!2}0.409
\\

& Night Light & \cellcolor{blue!3}\textbf{0.861}    & \cellcolor{green!3}\textbf{0.239} & \cellcolor{orange!2}\textbf{0.311} & \cellcolor{blue!3}\underline{0.850}  &     \cellcolor{green!3}\underline{0.244} & \cellcolor{orange!2}\underline{0.353} & \cellcolor{blue!3}0.531  &     \cellcolor{green!3}0.394 & \cellcolor{orange!2}0.629 & \cellcolor{blue!3}0.488  &     \cellcolor{green!3}0.429 & \cellcolor{orange!2}0.657 & \cellcolor{blue!3}0.369 &     \cellcolor{green!3}0.404 & \cellcolor{orange!2}0.728  & 
\cellcolor{blue!3}0.420 &     \cellcolor{green!3}0.457 & \cellcolor{orange!2}0.700 & 
\cellcolor{blue!3}0.668  &     \cellcolor{green!3}0.304 & \cellcolor{orange!2}0.414 & \cellcolor{blue!3}0.681   &     \cellcolor{green!3}0.301 & \cellcolor{orange!2}0.384
\\

& $PM_{2.5}$ & \cellcolor{blue!3}\textbf{0.921}    & \cellcolor{green!3}\textbf{0.064} & \cellcolor{orange!2}\textbf{0.160} & \cellcolor{blue!3}\underline{0.920} &     \cellcolor{green!3}\underline{0.065} & \cellcolor{orange!2}\underline{0.169} & \cellcolor{blue!3}0.641  &     \cellcolor{green!3}0.484 & \cellcolor{orange!2}0.954 & \cellcolor{blue!3}0.694  &     \cellcolor{green!3}0.306 & \cellcolor{orange!2}0.482 & \cellcolor{blue!3}0.398  &     \cellcolor{green!3}0.624 & \cellcolor{orange!2}0.845 & \cellcolor{blue!3}0.533 &     \cellcolor{green!3}0.556 & \cellcolor{orange!2}0.699 & \cellcolor{blue!3}0.791 &  \cellcolor{green!3}0.139 & \cellcolor{orange!2}0.271   
& \cellcolor{blue!3}0.809   &     \cellcolor{green!3}0.129 & \cellcolor{orange!2}0.234
\\

\midrule

\multirow{6}{*}{\rotatebox{90}{Shanghai}}
& Carbon & \cellcolor{blue!3}\textbf{0.915} &     \cellcolor{green!3}\textbf{0.157} & \cellcolor{orange!2}\textbf{0.290} & \cellcolor{blue!3}\underline{0.912}  &     \cellcolor{green!3}\underline{0.162} & \cellcolor{orange!2}\underline{0.312} & \cellcolor{blue!3}0.716  &     \cellcolor{green!3}0.392 & \cellcolor{orange!2}0.529 & \cellcolor{blue!3}0.688      & \cellcolor{green!3}0.413 & \cellcolor{orange!2}0.557 & \cellcolor{blue!3}0.298  &     \cellcolor{green!3}0.712 & \cellcolor{orange!2}0.914 & \cellcolor{blue!3}0.671  &     \cellcolor{green!3}0.426 & \cellcolor{orange!2}0.569 & \cellcolor{blue!3}0.696   &     \cellcolor{green!3}0.395 & \cellcolor{orange!2}0.563 & \cellcolor{blue!3}0.713   &     \cellcolor{green!3}0.388 & \cellcolor{orange!2}0.532
\\

& Population & \cellcolor{blue!3}\textbf{0.936}    & \cellcolor{green!3}\textbf{0.161} & \cellcolor{orange!2}\textbf{0.244} & \cellcolor{blue!3}\underline{0.928}  &     \cellcolor{green!3}\underline{0.172} & \cellcolor{orange!2}\underline{0.251} & \cellcolor{blue!3}0.593 &     \cellcolor{green!3}0.471 & \cellcolor{orange!2}0.607 & \cellcolor{blue!3}0.613  &     \cellcolor{green!3}0.456 & \cellcolor{orange!2}0.583 & \cellcolor{blue!3}0.315  &     \cellcolor{green!3}0.731 & \cellcolor{orange!2}0.959 & \cellcolor{blue!3}0.456  &     \cellcolor{green!3}0.557 & \cellcolor{orange!2}0.748 & \cellcolor{blue!3}0.617 &     \cellcolor{green!3}0.457& \cellcolor{orange!2}0.607 & \cellcolor{blue!3}0.686   &     \cellcolor{green!3}0.342 & \cellcolor{orange!2}0.498 
\\

& GDP & \cellcolor{blue!3}\textbf{0.778} &     \cellcolor{green!3}\textbf{0.323} & \cellcolor{orange!2}\textbf{0.468} & \cellcolor{blue!3}\underline{0.767} &     \cellcolor{green!3}\underline{0.331} & \cellcolor{orange!2}\underline{0.477} & \cellcolor{blue!3}0.330 &     \cellcolor{green!3}0.595 & \cellcolor{orange!2}0.816 & \cellcolor{blue!3}0.377 &     \cellcolor{green!3}0.553 & \cellcolor{orange!2}0.721 & \cellcolor{blue!3}0.294      & \cellcolor{green!3}0.767 & \cellcolor{orange!2}1.052 & \cellcolor{blue!3}0.326  &     \cellcolor{green!3}0.587 & \cellcolor{orange!2}0.807 & \cellcolor{blue!3}0.319 &     \cellcolor{green!3}0.611 & \cellcolor{orange!2}0.837 & \cellcolor{blue!3}0.331   &     \cellcolor{green!3}0.544 & \cellcolor{orange!2}0.811 
\\

& Night Light & \cellcolor{blue!3}\textbf{0.898} &     \cellcolor{green!3}\textbf{0.222} & \cellcolor{orange!2}\textbf{0.311} & \cellcolor{blue!3}\underline{0.891}  &     \cellcolor{green!3}\underline{0.234} & \cellcolor{orange!2}\underline{0.347} & \cellcolor{blue!3}0.457  &     \cellcolor{green!3}0.494 & \cellcolor{orange!2}0.667 & \cellcolor{blue!3}0.442  &     \cellcolor{green!3}0.517 & \cellcolor{orange!2}0.685 & \cellcolor{blue!3}0.308  &     \cellcolor{green!3}0.566& \cellcolor{orange!2}0.768 & \cellcolor{blue!3}0.387      & \cellcolor{green!3}0.511& \cellcolor{orange!2}0.709 & \cellcolor{blue!3}0.444 &     \cellcolor{green!3}0.571& \cellcolor{orange!2}0.756  & \cellcolor{blue!3}0.464   &     \cellcolor{green!3}0.517& \cellcolor{orange!2}0.729 
\\

& $PM_{2.5}$ & \cellcolor{blue!3}\textbf{0.866} &     \cellcolor{green!3}\textbf{0.120}& \cellcolor{orange!2}\textbf{0.379} & \cellcolor{blue!3}\underline{0.836}  &     \cellcolor{green!3}\underline{0.150}& \cellcolor{orange!2}\underline{0.394} & \cellcolor{blue!3}0.486  &     \cellcolor{green!3}0.497& \cellcolor{orange!2}0.654 & \cellcolor{blue!3}0.527 &     \cellcolor{green!3}0.398& \cellcolor{orange!2}0.592 & \cellcolor{blue!3}0.303 &     \cellcolor{green!3}0.617& \cellcolor{orange!2}0.895 & \cellcolor{blue!3}0.444 &     \cellcolor{green!3}0.518& \cellcolor{orange!2}0.774 & \cellcolor{blue!3}0.592 &     \cellcolor{green!3}0.342& \cellcolor{orange!2}0.554 & \cellcolor{blue!3}0.643   &     \cellcolor{green!3}0.291& \cellcolor{orange!2}0.542
\\

\midrule

\multirow{6}{*}{\rotatebox{90}{Guangzhou}}
& Carbon & \cellcolor{blue!3}\textbf{0.885}    & \cellcolor{green!3}\textbf{0.219}& \cellcolor{orange!2}\textbf{0.336} & \cellcolor{blue!3}\underline{0.884} &     \cellcolor{green!3}\underline{0.209}& \cellcolor{orange!2}\underline{0.371} & \cellcolor{blue!3}0.698  &     \cellcolor{green!3}0.385& \cellcolor{orange!2}0.514 & \cellcolor{blue!3}0.681  &     \cellcolor{green!3}0.497& \cellcolor{orange!2}0.529  & \cellcolor{blue!3}0.422  &     \cellcolor{green!3}0.708& \cellcolor{orange!2}0.708 & \cellcolor{blue!3}0.585  &     \cellcolor{green!3}0.444& \cellcolor{orange!2}0.603 & \cellcolor{blue!3}0.675   &     \cellcolor{green!3}0.415& \cellcolor{orange!2}0.503 & \cellcolor{blue!3}0.754   &     \cellcolor{green!3}0.370 & \cellcolor{orange!2}0.433
\\

& Population & \cellcolor{blue!3}\textbf{0.871}    & \cellcolor{green!3}\textbf{0.244}& \cellcolor{orange!2}\textbf{0.274} & \cellcolor{blue!3}\underline{0.855} &     \cellcolor{green!3}\underline{0.255}& \cellcolor{orange!2}\underline{0.299} & \cellcolor{blue!3}0.665  &     \cellcolor{green!3}0.441& \cellcolor{orange!2}0.481 & \cellcolor{blue!3}0.687  &     \cellcolor{green!3}0.433& \cellcolor{orange!2}0.485 & \cellcolor{blue!3}0.303  &     \cellcolor{green!3}0.954& \cellcolor{orange!2}0.972 & \cellcolor{blue!3}0.533  &     \cellcolor{green!3}0.567& \cellcolor{orange!2}0.687 & \cellcolor{blue!3}0.674   &     \cellcolor{green!3}0.471& \cellcolor{orange!2}0.483 & \cellcolor{blue!3}0.695  &     \cellcolor{green!3}0.433 & \cellcolor{orange!2}0.438
\\

& GDP & \cellcolor{blue!3}\textbf{0.715} &     \cellcolor{green!3}\textbf{0.371}& \cellcolor{orange!2}\textbf{0.532} & \cellcolor{blue!3}\underline{0.712}  &     \cellcolor{green!3}\underline{0.366}& \cellcolor{orange!2}\underline{0.574}  & \cellcolor{blue!3}0.436  &     \cellcolor{green!3}0.541& \cellcolor{orange!2}0.764 & \cellcolor{blue!3}0.439  &     \cellcolor{green!3}0.533 & \cellcolor{orange!2}0.699& \cellcolor{blue!3}0.282      & \cellcolor{green!3}0.897& \cellcolor{orange!2}1.264 & \cellcolor{blue!3}0.440  &     \cellcolor{green!3}0.546& \cellcolor{orange!2}0.762 & \cellcolor{blue!3}0.411   &     \cellcolor{green!3}0.625& \cellcolor{orange!2}0.748 & \cellcolor{blue!3}0.476   &     \cellcolor{green!3}0.611& \cellcolor{orange!2}0.763
\\

& Night Light & \cellcolor{blue!3}\textbf{0.871} &     \cellcolor{green!3}\textbf{0.234}& \cellcolor{orange!2}\textbf{0.378} & \cellcolor{blue!3}\underline{0.854}  &     \cellcolor{green!3}\underline{0.248}& \cellcolor{orange!2}\underline{0.391} & \cellcolor{blue!3}0.577  &     \cellcolor{green!3}0.418& \cellcolor{orange!2}0.573  & \cellcolor{blue!3}0.574  &     \cellcolor{green!3}0.415& \cellcolor{orange!2}0.581 & \cellcolor{blue!3}0.435  &     \cellcolor{green!3}0.433& \cellcolor{orange!2}0.627 & \cellcolor{blue!3}0.483      & \cellcolor{green!3}0.478& \cellcolor{orange!2}0.633 & \cellcolor{blue!3}0.442  &     \cellcolor{green!3}0.451& \cellcolor{orange!2}0.604 & \cellcolor{blue!3}0.476   &     \cellcolor{green!3}0.402 & \cellcolor{orange!2}0.598
\\

& $PM_{2.5}$ & \cellcolor{blue!3}\textbf{0.833}    & \cellcolor{green!3}\textbf{0.133}& \cellcolor{orange!2}\textbf{0.403} & \cellcolor{blue!3}\underline{0.822} &     \cellcolor{green!3}\underline{0.158}& \cellcolor{orange!2}\underline{0.414} & \cellcolor{blue!3}0.638  &     \cellcolor{green!3}0.462& \cellcolor{orange!2}0.624 & \cellcolor{blue!3}0.652 &     \cellcolor{green!3}0.461& \cellcolor{orange!2}0.597 & \cellcolor{blue!3}0.315  &     \cellcolor{green!3}0.542& \cellcolor{orange!2}0.741 & \cellcolor{blue!3}0.56 &     \cellcolor{green!3}0.514& \cellcolor{orange!2}0.694 & \cellcolor{blue!3}0.671  &     \cellcolor{green!3}0.394 & \cellcolor{orange!2}0.559& \cellcolor{blue!3}0.696   &     \cellcolor{green!3}0.380& \cellcolor{orange!2}0.503
\\

\midrule

\multirow{6}{*}{\rotatebox{90}{Shenzhen}}
& Carbon & \cellcolor{blue!3}\textbf{0.926}    & \cellcolor{green!3}\textbf{0.128}& \cellcolor{orange!2}\textbf{0.290} & \cellcolor{blue!3}\underline{0.912} &     \cellcolor{green!3}\underline{0.162}& \cellcolor{orange!2}\underline{0.304} & \cellcolor{blue!3}0.659  &     \cellcolor{green!3}0.418& \cellcolor{orange!2}0.568 & \cellcolor{blue!3}0.647  &     \cellcolor{green!3}0.431& \cellcolor{orange!2}0.581  & \cellcolor{blue!3}0.257  &     \cellcolor{green!3}0.683& \cellcolor{orange!2}0.816 & \cellcolor{blue!3}0.562  &     \cellcolor{green!3}0.483& \cellcolor{orange!2}0.571 & \cellcolor{blue!3}0.241  &     \cellcolor{green!3}0.577& \cellcolor{orange!2}0.726  & \cellcolor{blue!3}0.189   &     \cellcolor{green!3}0.634& \cellcolor{orange!2}0.887 
\\

& Population & \cellcolor{blue!3}\textbf{0.892}    & \cellcolor{green!3}\textbf{0.165}& \cellcolor{orange!2}\textbf{0.244} & \cellcolor{blue!3}\underline{0.879}  &     \cellcolor{green!3}\underline{0.173}& \cellcolor{orange!2}\underline{0.261} & \cellcolor{blue!3}0.790  &     \cellcolor{green!3}0.343& \cellcolor{orange!2}0.448 & \cellcolor{blue!3}0.797  &     \cellcolor{green!3}0.314& \cellcolor{orange!2}0.390 & \cellcolor{blue!3}0.311 &     \cellcolor{green!3}0.758& \cellcolor{orange!2}0.892 & \cellcolor{blue!3}0.527   &     \cellcolor{green!3}0.592& \cellcolor{orange!2}0.610 & \cellcolor{blue!3}0.719 &     \cellcolor{green!3}0.354& \cellcolor{orange!2}0.455 & \cellcolor{blue!3}0.775   &     \cellcolor{green!3}0.342 & \cellcolor{orange!2}0.421
\\

& GDP & \cellcolor{blue!3}\textbf{0.798}    & \cellcolor{green!3}\textbf{0.297}& \cellcolor{orange!2}\textbf{0.468} & \cellcolor{blue!3}\underline{0.767} &     \cellcolor{green!3}\underline{0.331}& \cellcolor{orange!2}\underline{0.489} & \cellcolor{blue!3}0.532  &     \cellcolor{green!3}0.448& \cellcolor{orange!2}0.676 & \cellcolor{blue!3}0.517  &     \cellcolor{green!3}0.455& \cellcolor{orange!2}0.699 & \cellcolor{blue!3}0.307  &     \cellcolor{green!3}0.895& \cellcolor{orange!2}1.003 & \cellcolor{blue!3}0.508  &     \cellcolor{green!3}0.464& \cellcolor{orange!2}0.693& \cellcolor{blue!3}0.534  &     \cellcolor{green!3}0.417 & \cellcolor{orange!2}0.625& \cellcolor{blue!3}0.553   &     \cellcolor{green!3}0.404& \cellcolor{orange!2}0.623
\\

& Night Light & \cellcolor{blue!3}\textbf{0.942} &     \cellcolor{green!3}\textbf{0.149}& \cellcolor{orange!2}\textbf{0.245} & \cellcolor{blue!3}\underline{0.939}  &     \cellcolor{green!3}\underline{0.155}& \cellcolor{orange!2}\underline{0.268} & \cellcolor{blue!3} 0.457  &     \cellcolor{green!3}0.459& \cellcolor{orange!2}0.667 & \cellcolor{blue!3} 0.445 &     \cellcolor{green!3}0.488& \cellcolor{orange!2}0.682 & \cellcolor{blue!3}0.454  &     \cellcolor{green!3}0.358& \cellcolor{orange!2}0.588  & \cellcolor{blue!3}0.387      & \cellcolor{green!3}0.511& \cellcolor{orange!2}0.709 & \cellcolor{blue!3}0.413  &     \cellcolor{green!3}0.513& \cellcolor{orange!2}0.719 & \cellcolor{blue!3}0.396   &     \cellcolor{green!3}0.528& \cellcolor{orange!2}0.758  \\

& $PM_{2.5}$ & \cellcolor{blue!3}\textbf{0.906}    & \cellcolor{green!3}\textbf{0.116}& \cellcolor{orange!2}\textbf{0.298} & \cellcolor{blue!3}\underline{0.905}  &     \cellcolor{green!3}\underline{0.117}& \cellcolor{orange!2}\underline{0.331} & \cellcolor{blue!3}0.566  &     \cellcolor{green!3}0.494& \cellcolor{orange!2}0.598 & \cellcolor{blue!3}0.597 &    \cellcolor{green!3}0.451& \cellcolor{orange!2}0.491  & \cellcolor{blue!3}0.323  &     \cellcolor{green!3}0.613& \cellcolor{orange!2}0.883 & \cellcolor{blue!3}0.430 &     \cellcolor{green!3}0.586 & \cellcolor{orange!2}0.762& \cellcolor{blue!3}0.653 &     \cellcolor{green!3}0.398& \cellcolor{orange!2}0.426 & \cellcolor{blue!3}0.749   &     \cellcolor{green!3}0.345& \cellcolor{orange!2}0.391
\\

\bottomrule
\end{tabular}

}
\vspace{-0.5em}
\end{small}
\end{table*}

\subsubsection{Baselines.} We compare our approach with several baselines from the following two areas: 1) Traditional spatial interpolation. 2) region representation. We first introduce the following two approaches for traditional interpolation:  
\begin{itemize}[leftmargin=*]
    \item \textbf{IDW (Inverse Distance Weighting)} \citep{zimmerman1999experimental} A simple and widely used spatial interpolation method that estimates values at unsampled locations by weighting the values of nearby sampled points inversely by their distance to the target location. It assumes that the influence of a point decreases monotonously with distance, making it effective when data points are dense and evenly distributed.
    \item \textbf{Universal Kriging} \citep{brus2007optimization} A geostatistical interpolation technique that considers both the distance and the spatial correlation between data points. It incorporates a trend component into the interpolation model, allowing for more accurate predictions in the presence of spatially varying trends. 
\end{itemize}
Then we compare our methods with four region representation approaches based on multi-modal geospatial data for socioeconomic indicator prediction as follows:
\begin{itemize}[leftmargin=*]
    \item \textbf{UrbanCLIP}: \citep{yan2023urban} A model incorporating a multi-modal Large Language Model (LLM) to enhance the encoding of satellite imagery. This model generates descriptive texts for satellite images using the LLM and then fuses these texts with the images through an image-text contrastive learning-based approach to capture the complexity and diversity of geospatial areas.
    \item \textbf{PG-SimCLR} \citep{xi2022beyond}: A contrastive learning framework that introduces societal information (i.e., POI) into geospatial region representation learning from satellite imagery.
    \item \textbf{GeoStructural} \citep{li2022predicting}: The graph-based framework profiles geospatial regions by utilizing street segments as graph structure for adaptively fusing features from multi-level satellite and street-view images. For convenience, we refer to this method as GeoStructural.
    \item \textbf{UrbanVLP} \citep{hao2024urbanvlp}: An region representation method based on contrastive learning, which incorporates satellite imagery, street-view images, and spatial position structure. This method is further enhanced by incorporating a Large Language Model (LLM) and GeoCLIP \citep{vivanco2024geoclip}, resulting in improved robustness and performance.
\end{itemize}

To evaluate the inference approaches of our method, we introduce two different variants of \model:

\begin{itemize}[leftmargin=*]
\item \textbf{GeoHG} This variant employs an end-to-end inference approach, directly training the overall framework on each dataset, similar to most spatial interpolation methods, to achieve optimal inference performance.
\item \textbf{GeoHG-SSL} This variant follows the conventional setting for region representation methods, first generating a universal region embedding matrix during the graph's self-supervised pre-training stage. The model then only requires fine-tuning of the MLP layers for regression on various tasks.
\end{itemize}


\subsubsection{Metrics \& Implementation.}
We employ mean absolute error (MAE), rooted mean squared error (RMSE), and coefficient of determination ($R^2$) as evaluation metrics. \model is trained using Adam optimizer with a learning rate of $2e^{-3}$.  Higher $R^2$ and lower RMSE, MAE means better performance. For the hyperedge gate $\theta_{Env}$ and $\theta_{Soc}$, we conduct grid searches over \{0.2, 0.4, 0.6, 0.8\} and \{0.3, 0.6, 0.9, 1.2, 1.5\} respectively. For the number of layers of the graph convolution block, we test it from 1 to 3. The execution time required for training the \model is $\sim$\textit{3 minutes} per task.
We introduce the best hyperparameter configurations for each task as below:
\begin{itemize}[leftmargin=*]
    \item For the Carbon dataset, the hypergate $\theta_{Env}$ is set as 0.6 while $\theta_{Soc}$ is set as 0.9. The number of layers of the graph convolution block is 3, and the dimension of the hidden layer is 64.
    \item For the Population dataset, the hypergate $\theta_{Env}$ is set as 0.2 while $\theta_{Soc}$ is set as 0.9. The number of layers of the graph convolution block is 3, and the dimension of the hidden layer is 64.
    \item For the GDP dataset, the hypergate $\theta_{Env}$ is set as 0.4 while $\theta_{Soc}$ is set as 1.2. The number of layers of the graph convolution block is 3, and the dimension of the hidden layer is 64.
    \item For the Night Light dataset, the hypergate $\theta_{Env}$ is set as 0.2 while $\theta_{Soc}$ is set as 0.9. The number of layers of the graph convolution block is 3, and the dimension of the hidden layer is 64.
    \item For the $PM_{2.5}$ dataset, the hypergate $\theta_{Env}$ is set as 0.8 while $\theta_{Soc}$ is set as 0.6. The number of layers of the graph convolution block is 3, and the dimension of the hidden layer is 64.
    
\end{itemize}

\subsection{Comparison with State-of-the-Art Methods}

We conduct an empirical assessment of various models across the four datasets, and the results of each model are detailed in Table \ref{tab:results_overall}. From these tables, we have two key findings: \textbf{1)}Both \textbf{\model} and \textbf{\model-SSL} outperform all competing baselines over the 5 datasets for 4 cities. For instance, \model surpasses the previous SOTA performance in Beijing, achieving $R_2$ improvements of \underline{+16.7\%, +14.9\%, +6.1\%, +37\%, +33\%} in Carbon, Population, GDP, Night Light and $PM_{2.5}$ tasks respectively. Consistent trends are observed across other cities, underscoring \model's stable accuracy. \textbf{2)} Most socioeconomic indicator prediction methods perform even worse than traditional spatial interpolation methods on indicator inference tasks. GeoHG captures both spatial geometry and the underlying socioeconomic dynamics of geospace, resulting in lower data reliance compared to existing methods.

\begin{table}[!h]
\small
  \centering
  \caption{Large-scale Interpolation Result.}
  \vspace{-1em}
  \resizebox{1.0\columnwidth}{!}{
    \begin{tabular}{cccccc}
    \toprule
        Metric &GeoHG & IDW & UK & UrbanCLIP   & GeoStructural \\
    \midrule
    $R^2$ & \textbf{0.813}  & 0.693  & 0.735  & 0.303  & 0.594 \\
    MAE   & \textbf{0.298} & 0.558 & 0.423 & 0.844 & 0.732 \\
    RMSE  & \textbf{0.375}  & 0.746 & 0.635 & 0.964 & 0.856 \\
    \bottomrule
    \end{tabular}%
  \label{tab:large}%
  }
\end{table}%

\begin{figure}[!b]
\vspace{-1em}
    \label{sec:fewshot2}
    \centering
    \includegraphics[width=0.95\linewidth]{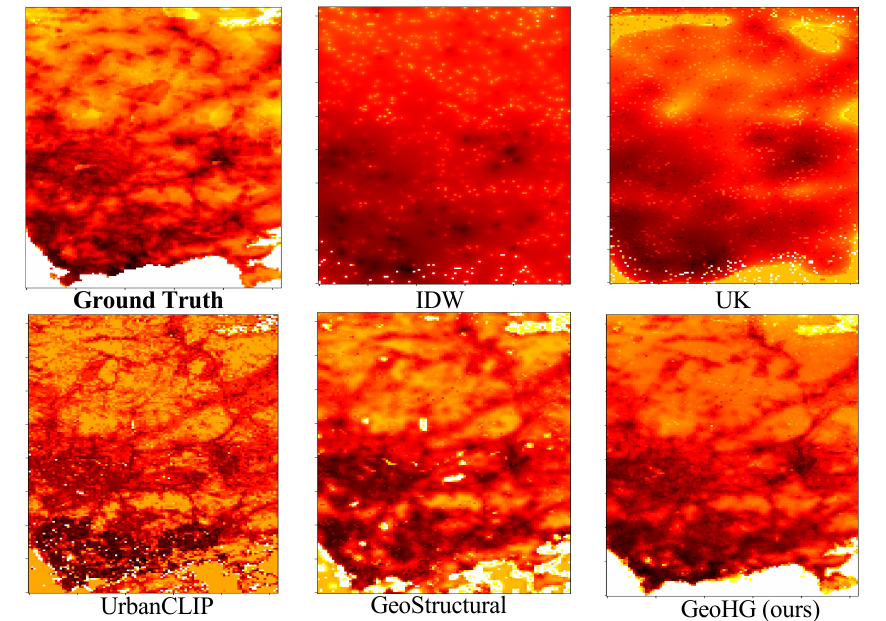}
    \vspace{-1em}
    \caption{Visualization of Large-scale Inference Results. }
    \label{fig:fewshot2}
\end{figure}

\subsection{Large-scale Interpolation}
To better evaluate the advantages of GeoHG's non-continuous inference on large-scale geospace, we assess the model against baseline methods for population inference in an expanded area of Shenzhen. This evaluation uses only \uline{5\%} of the available data (863 points) to infer the population of the remaining \textbf{16,401} $km^2$ area, as illustrated in Figure \ref{fig:fewshot2} and Table \ref{tab:large}. The model accurately captures the true distribution and substantially outperforms traditional spatial interpolation methods such as Inverse Distance Weighting (IDW) and Kriging (UK), which show significant deviations. The variation in geospace between continent, sea, mountains, and coastline is clearly represented with sharp outlines, leading to optimal accuracy. This outcome demonstrates that traditional methods, which rely solely on spatial relationships for interpolation, do not adequately model the complex socio-environmental dependencies essential for characterizing the distribution of real-world socioeconomic indicators. In contrast, the effective joint modeling of geospace by GeoHG allows for more accurate inference of indicators beyond single spatial geometric dimension.

    

\begin{figure*}[!t]
    \centering
    \includegraphics[width=0.75\linewidth]{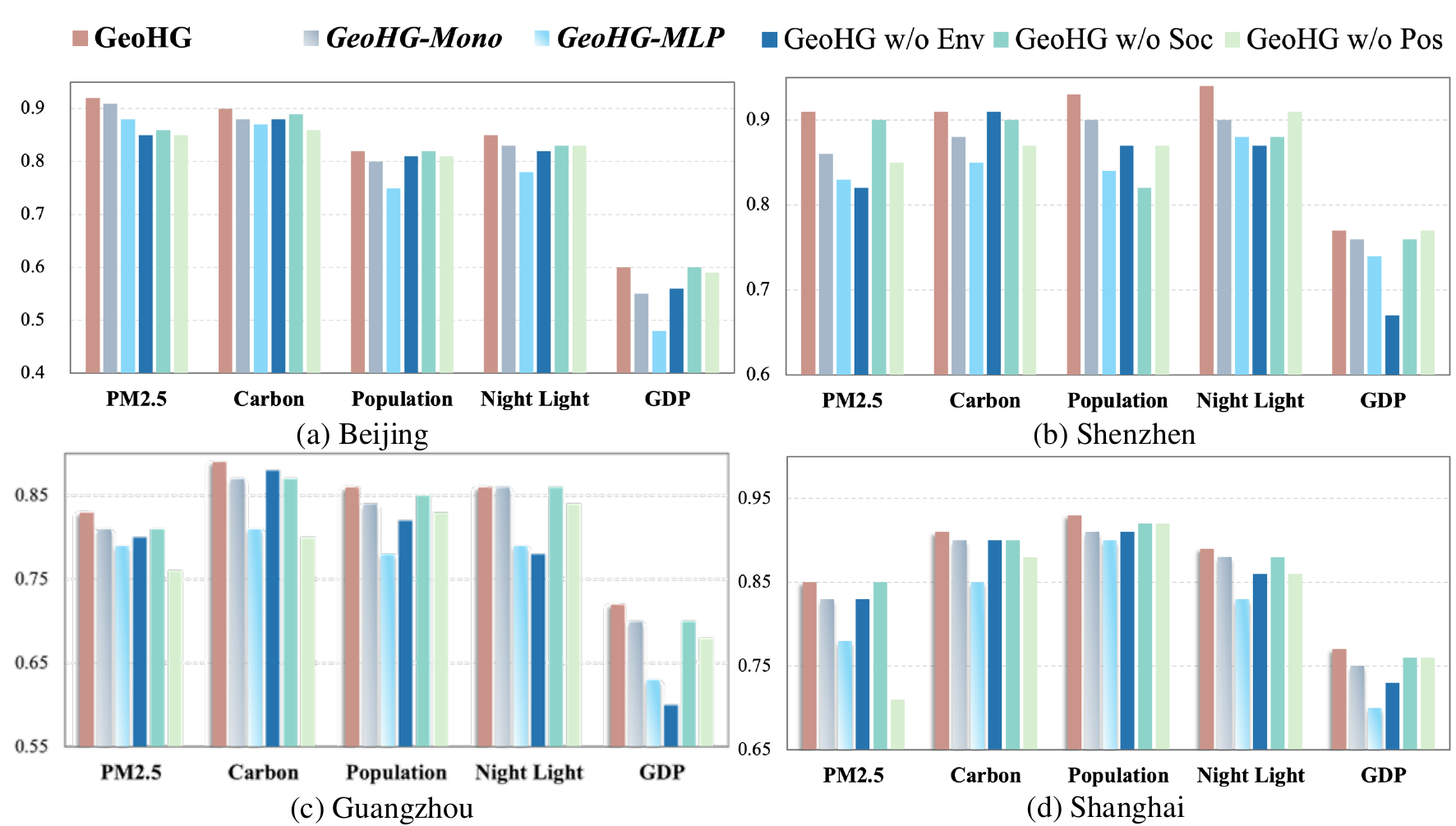}
    \vspace{-1em}
    \caption{Results of Ablation Study on $R^2$ Metric.}
    \vspace{-1em}
    \label{fig:ablation}
\end{figure*}

\subsection{Ablation Study}

\textbf{Effects of Core Components.} To examine the effectiveness of each core module, we conducted an ablation study based on the following variants: a) \textbf{w/o Env}, which excludes environmental information from the satellite imagery for embedding; b) \textbf{w/o Soc}, which omits POI entities for embedding; and c) \textbf{w/o Pos}, which does not utilize the corresponding location information of the regions. The $R^2$ results for five tasks in four cities are shown in Figure~\ref{fig:ablation}. We observe that removing location information significantly degrades performance in all tasks. Additionally, environmental features have a substantial impact on model accuracy for the PM2.5, Night Light, and GDP tasks. These results align with the intrinsic distribution characteristics of the task datasets. For instance, PM2.5 and night light data are primarily related to environmental elements such as vegetation and infrastructure. Conversely, data for carbon estimation is mainly derived from societal activity metrics. The analysis of the four cities reveals distinct distribution patterns across the five key indicators, highlighting the heterogeneity in geospatial characteristics. However, the robust performance of our model, which is constructed through the synthesis of three distinct perspectives, remains evident.

\noindent \textbf{Effect of Graph Construction.} Our framework utilizes heterogeneous graph and hyperedges between regions to reflect the geospatial high-order relations. To validate the effectiveness of high-order relation representation, we compare \model against two variants: \textbf{\model-MLP} which discards graph structure and relies only on intra-region feature representation, employing MLP for regression; \textbf{\model-Mono} which keeps edges in adjacent regions while discarding hyperedges for high-order relations. The results presented in Figure~\ref{fig:ablation} indicated that our \model significantly benefits from efficient representation of complex mixed-order dependiencies in geospace, thereby resulting in enhanced performance.

\subsection{ Interpretation Analysis}
\begin{figure}
    \centering
    \includegraphics[width=1.0\linewidth]{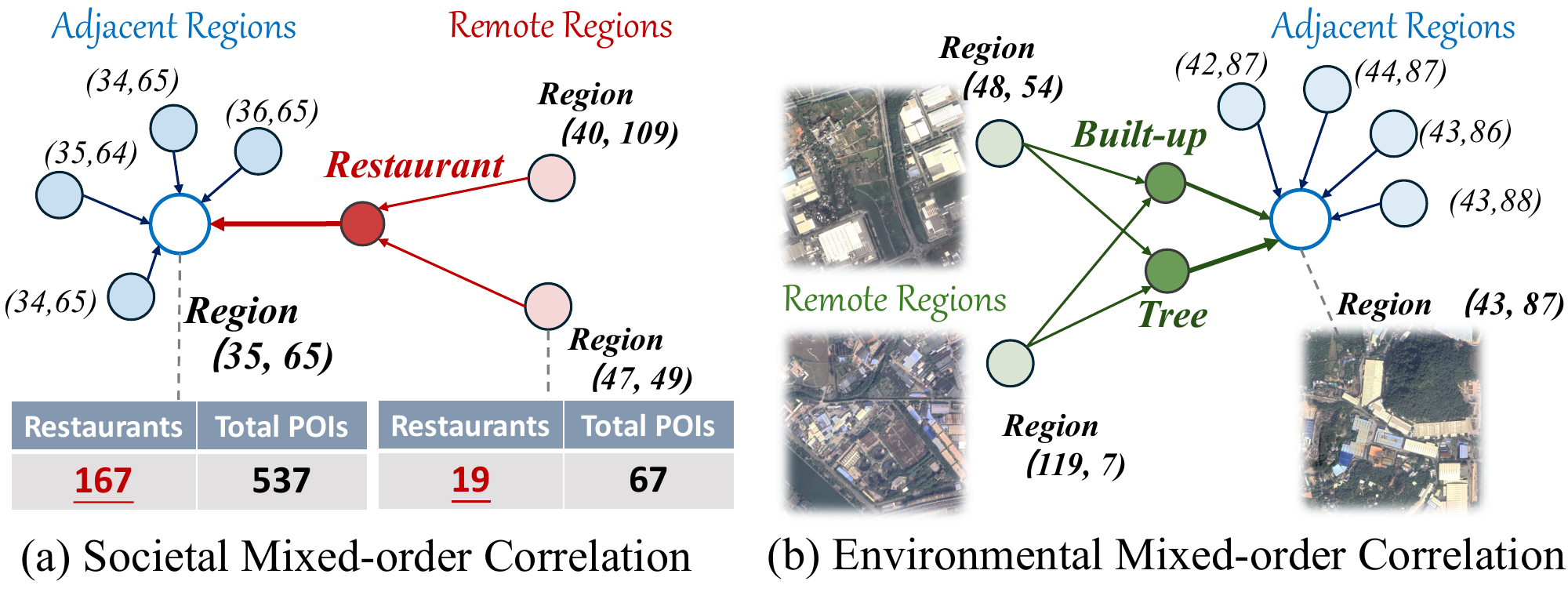}
    \vspace{-1.5em}
    \caption{Visualization of Geospatial Correlations learned by \model. $(x,y)$ represents the numbering in the latitude and longitude.}
    \vspace{-1em}
    \label{fig:case1}
\end{figure}

To validate \model's efficacy in capturing higher-order non-continuous geospatial relationships, we employ GNNExplainer~\citep{ying2019gnnexplainer}, a model-agnostic technique for interpreting GNN predictions by identifying crucial subgraph structures and features. Specifically, for the task of inferring the unknown carbon emission of the region \#(35,65) in Guangzhou, we utilized GNNExplainer to extract the top 10 most influential edges, as shown in Table~\ref{tab:gnnexplainer-societal}. Similar experimental results for region \#(43,87) are presented in Table~\ref{tab:gnnexplainer-env}. Based on these tables and the corresponding Figure \ref{fig:case1}, we can derive the following key findings:

\begin{table}[b!]
\centering
\caption{Top 10 important edges for carbon emission predictions for region \#(35,65) in Guangzhou. The underlined font is the regional nodes adjacent to the target region, and the bold font refers to the distant region nodes.}
\label{tab:gnnexplainer-societal}
\resizebox{1.0\columnwidth}{!}{
\begin{tabular}{ccccc}
\toprule
\makecell[c]{Source Node\\Type} & \makecell[c]{Source Node\\Name}  & \makecell[c]{Target Node\\Type} & \makecell[c]{Target Node\\Name}  & Importance \\
\midrule
region           & (35,65)           & society          & Food and Beverage & 0.534    \\
region           & (35,65)           & society          & Shopping Mall     & 0.511      \\
region           & (35,65)           & environment      & Built-up          & 0.522      \\
region           & (35,65)           & region           & \underline{(34,65)}  & 0.526      \\
region           & (35,65)           & region           & \underline{(36,65)}           & 0.519      \\
region           & (35,65)           & region           & \underline{(35,64)}           & 0.522      \\
region           & (35,65)           & region           & \underline{(34,64)}           & 0.520      \\
society          & Food and Beverage & region           & \textbf{(40,109)}          & 0.495      \\
society          & Food and Beverage & region           & \textbf{(47,49)}           & 0.495\\  
\bottomrule
\end{tabular}
}
\end{table}
\begin{table}[h!]
\centering
\caption{Top 10 important edges for carbon emission predictions for region \#(43,87) in Guangzhou. The underlined font is the regional nodes adjacent to the target region, and the bold font refers to the distant region nodes.}
\label{tab:gnnexplainer-env}
\resizebox{1.0\columnwidth}{!}{
\begin{tabular}{ccccc}
\toprule
\makecell[c]{Source Node\\Type} & \makecell[c]{Source Node\\Name}  & \makecell[c]{Target Node\\Type} & \makecell[c]{Target Node\\Name}  & Importance \\
\midrule
region           & (43,87)           & society          & Food and Beverage & 0.496  \\
region           & (43,87)           & environment          & Built-up     & 0.495      \\
region           & (43,87)           & environment      & Tree          & 0.519      \\
region           & (43,87)           & region           & \underline{(42,87)}  & 0.531      \\
region           & (43,87)           & region           & \underline{(44,87)}           & 0.526      \\
region           & (43,87)           & region           & \underline{(43,86)}           & 0.524      \\
region           & (43,87)           & region           & \underline{(43,88)}           & 0.520      \\
environment          & Built-up & region           & \textbf{(48,54)}          & 0.497      \\
environment         & Tree & region           & \textbf{(119,7)}           & 0.495\\  
\bottomrule
\end{tabular}
}
\end{table}

\begin{enumerate}[label=\arabic*), leftmargin=*]
    \item The inference for the target region exhibits strong associations with heterogeneous societal or environmental entity nodes such as "Food and Beverage" and "Shopping Mall," which are typically carbon-intensive due to factors like energy usage (cooking, refrigeration), transportation (goods/personnel movement), and packaging consumption.
    \item Carbon emissions from adjacent regions also emerge as highly useful values, an intuitive pattern resulting from spatial proximity and potential environmental spillovers between neighboring areas.
    \item Our graph structure effectively connects the target region to relatively distant areas like \#(40, 109) and \#(47, 49). Despite geographical separation, these regions share similar social contexts, being linked to the "Food and Beverage" node, thereby providing relevant emission patterns to inform the prediction.
\end{enumerate}

These interpretable insights validate GeoHG's effectiveness in capturing the intricate high-order dependencies between environmental conditions, urbanization factors, economic activities, and their synergistic impact on carbon footprints across regions. These characteristics enable our model to extract more complex information from geospatial data for indicator inference.

\subsection{Qualitative Demonstration}

\begin{figure}[h!]
    \centering
    \includegraphics[width=1.0\linewidth]{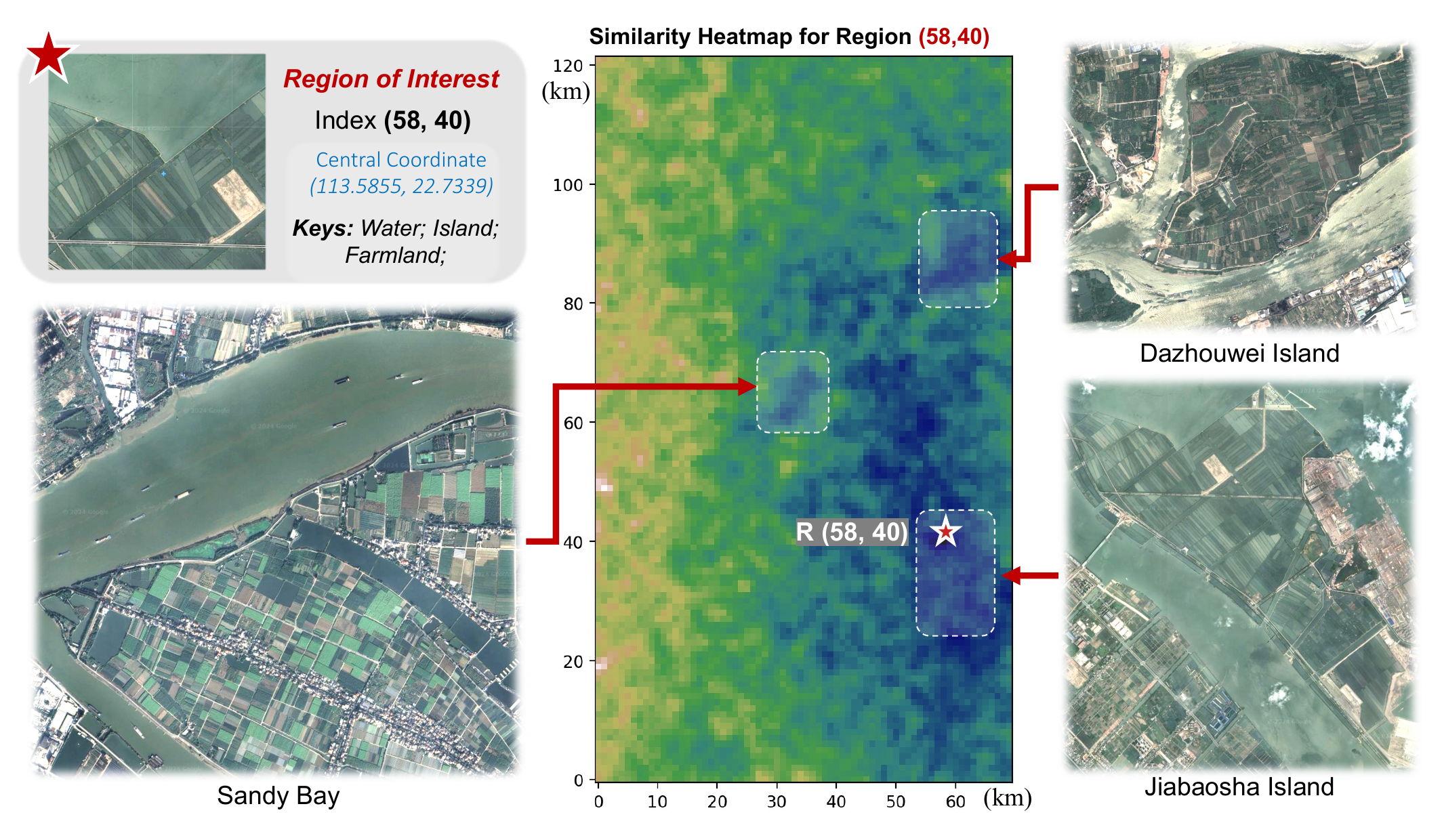}
    \includegraphics[width=1.0\linewidth]{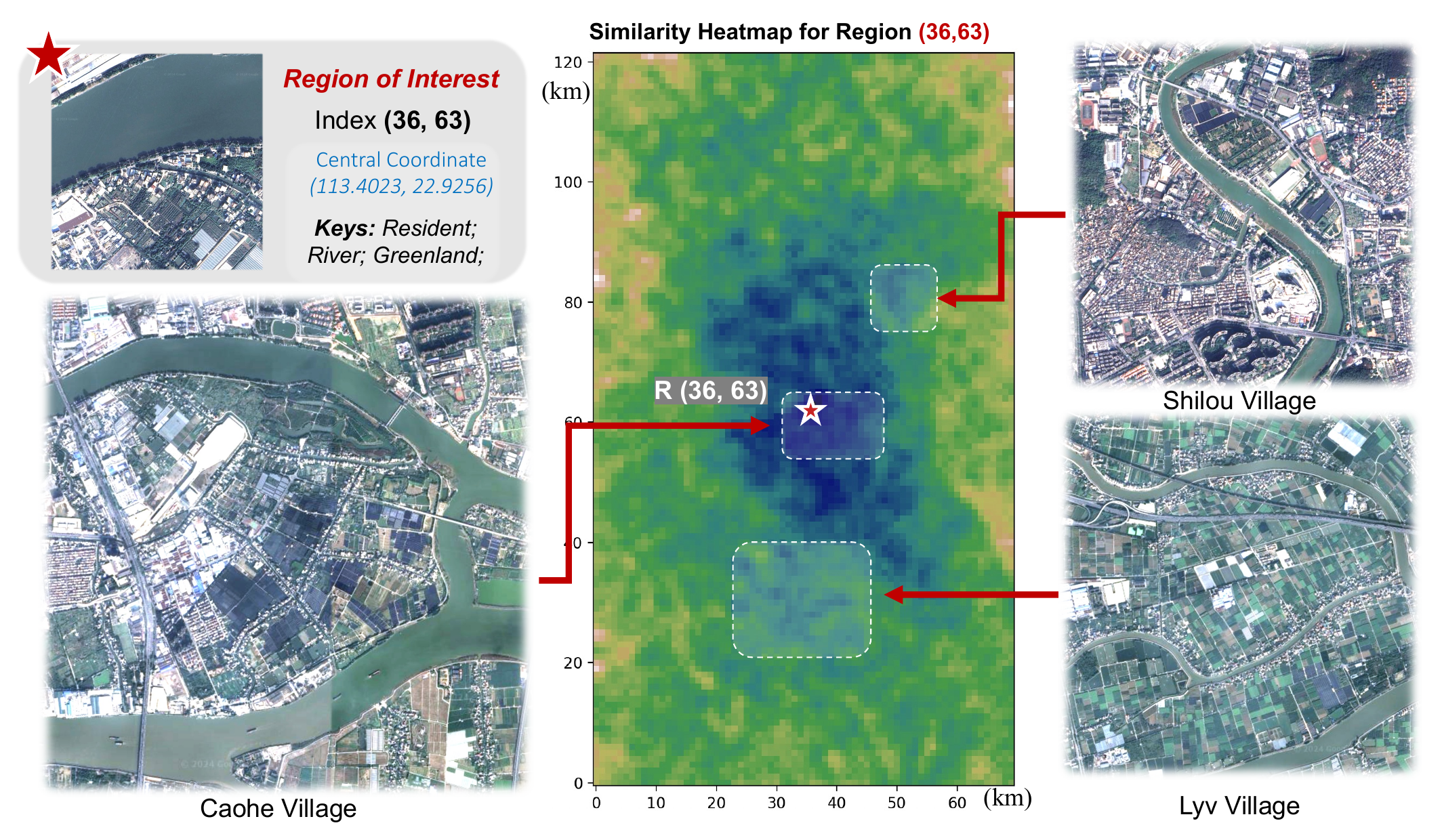}
    \vspace{-2em}
    \caption{Regions' similarity in GeoHG of Guangzhou. \textbf{Darker} indicate higher similarity with the region of interest.}
    \vspace{-0.5em}
    \label{fig:simi_visual}
\end{figure}

To qualitatively assess the effectiveness of GeoHG in modeling non-continuity and heterogeneity of geospace, we compute the cosine similarity between different regions using embedding matrix of space generated by GeoHG-SSL. The regional dependency distribution is illustrated in Figure \ref{fig:simi_visual}. The cosine similarity \(S_{C}\) between the embedding \(E_A\) of region A and any other region's embedding \(E_B\) is calculated as follows:

\[
S_{C}(E_A, E_B) = \frac{\mathbf{E_A} \cdot \mathbf{E_B}}{\|\mathbf{E_A}\|\|\mathbf{E_B}\|} = \frac{\sum_{i=1}^{n} E_{A i} E_{B i}}{\sqrt{\sum_{i=1}^{n} E_{A i}^{2}} \cdot \sqrt{\sum_{i=1}^{n} E_{B i}^{2}}}
\]

where \(E_{A i}\) and \(E_{B i}\) are the \(i_{th}\) components of \(E_A\) and \(E_B\), respectively.

The visualization results clearly show non-continuous spatial characters chaptered by our model. For Region (40,58), a farmland area near water on Jiabaosha Island, our model identifies nearby regions with higher similarity. It also connects to remote islands, such as Dazhouwei Island and Sandy Bay, indicating similar environmental functions. In contrast, the mountainous industrial area in western Guangzhou shows lower similarity. For Region (36,63), an urban residential area named Caohe Village, the heatmap demonstrates connections to other similar residential areas like Shilou Village and Lyv Village, which are adjacent to flowing rivers but not connected from normal spatial dimension.

\subsection{Robustness Testing on Data Efficiency}

To validate the robustness of GeoHG for inference with extremely limited data, we evaluate the model's performance under masked ratios of \underline{1\%, 5\%, 10\%, and 15\%}, with results shown in Table~\ref{tab:fewshot}. The results demonstrate the model's strong data efficiency; with only 5\% of the data, the model outperforms previous state-of-the-art socioeconomic indicator prediction models, such as UrbanVLP~\citep{hao2024urbanvlp} and GeoStructural~\citep{li2022predicting}, even though these models were trained on the entire training dataset across all tasks as described in their original papers.

\begin{table}[htbp]
\small
  \centering
  \caption{Data Efficiency of \model in indicator inference. We take the average of $R^2$ results in 4 cities.}
  \vspace{-1em}
  \resizebox{1.0\columnwidth}{!}{
    \begin{tabular}{cccccc}
    \toprule
        Masked Ratio  & Carbon & Population & GDP   & Light & PM2.5 \\
    \midrule
    99\% & 0.720  & 0.713  & 0.535  & 0.703  & 0.694 \\
    95\%   & 0.835 & 0.818 & 0.623 & 0.844 & 0.832 \\
    90\%  & 0.850  & 0.846 & 0.635 & 0.864 & 0.856 \\
    80\%  & 0.877 & 0.861 & 0.724 & 0.876 & 0.867 \\
    \bottomrule
    \end{tabular}%
  \label{tab:fewshot}%
  }
\end{table}%

\section{Related Work}
\textbf{Deep Learning for Geospace Embedding.}
Numerous prior studies \citep{zhang2021multi,zhou2023heterogeneous,huang2023comprehensive,wang2020urban2vec,li2022predicting,yang2024graphagent} have tackled spatial embedding in data-driven solutions. For example, \citet{zhang2021multi} used a multi-view graph representation that integrates POI data and mobility patterns, while \citet{zhou2023heterogeneous} applied a prompt learning technique combining both POI and mobility data. Recent models like PG-SimCLR \citep{xi2022beyond} and UrbanCLIP \citep{yan2023urban} have succeeded in profiling urban regions through satellite images. However, these models still struggle with indicator inference from limited samples and do not sufficiently explore the joint representation of geospace from geometric, environmental, and societal perspectives.

\noindent
\textbf{Socioeconomic Indicator Inference.} Socioeconomic indicator inference is based on spatial interpolation methods, including Inverse Distance Weighting (IDW)~\citep{zimmerman1999experimental,lu2008adaptive} and Kriging~\citep{brus2007optimization}, which are commonly used for this purpose. IDW is a straightforward method that weights sample points inversely by distance, but it does not consider spatial correlation and assumes a monotonous decrease in influence with distance~\citep{mitas1999spatial}. In contrast, Kriging is a more complex method that incorporates both distance and spatial correlation, utilizing variograms and covariance functions to minimize prediction errors and provide uncertainty estimates. However, Kriging is computationally intensive for large spatial domains and relies on assumptions of spatial stationarity and isotropy~\citep{leal2025interpreting}. These widely-used interpolation methods are built on the assumptions of spatial continuity and isotropy.

\noindent
\textbf{Multimodal Learning in GeoAI.}
Multimodal learning integrates data from various modalities to enhance model performance across tasks \citep{baltruvsaitis2018multimodal,guo2019deep,gao2020survey}. Traditional geospatial embedding learning primarily relies on task-specific supervised methods \citep{yuan2012discovering,zheng2015methodologies}, which are limited by their dependence on domain expertise and their inability to adapt to new tasks \citep{zou2024deep,guo2021learning}.
Recent studies \citep{jenkins2019unsupervised, zhang2021multi} have aimed at learning general geospace embeddings from a multimodal perspective for unified region embedding. However, efficiently integrating multimodal data from complex geospaces while maintaining generalizability remains a significant challenge \citep{shang2020introduction,zou2024deep,rao2024next,shang2020deep}. Conventional vision encoders, including multimodal large language models, struggle to capture geo-semantic details of satellite images, and often overlook the social impact of POIs in geospace. Furthermore, popular modality fusion techniques in multimodal learning, such as CLIPs \citep{hao2024urbanvlp,yan2023urban}, fail to adequately address the cross-modal interactions between geo-entities within geospace.In this research, we propose an integrated geospace modeling learning framework based on heterogeneous graph architecture, inspired by successful semantic segmentation of satellite imagery \citep{zanaga2021esa}. This framework merges visual entities and POI attributes while considering mixed-order geospatial relationships among entities and regions.
\section{Discussion on Social Impacts}
\label{discussion}
Efficient socioeconomic indicator inference holds considerable benefits for the broader geographic and spatial computing communities. By comprehensively representing intra-region features and inter-region correlations, our proposed \model framework exhibits significant potential to effect meaningful change across various domains, especially within the realms of smart cities and geoscience. The interoperability and efficiency provided by \model facilitate a deeper understanding of complex geospace and the underlying mixed-order correlations inherent in the space. This comprehensive representation of regional spatial regions empowers stakeholders to monitor cities and environments more effectively, thereby making informed decisions that ultimately enhance individual and communal quality of life while fostering more resilient and sustainable environments.

Moreover, the remarkable data efficiency of \model enables the community to investigate regional fine-grained climates with limited resources. For instance, in our experiment, we showcase its exceptional performance in predicting fine-grained ($1km \times 1km$) indicators for a large area ($16,401 km^2$) using only 863 monitoring points. As human society progresses and the global environment changes, regional extreme climates, such as urban heat islands and local air pollution, continue to pose economic, environmental, and health challenges. Therefore, effective regional climate monitoring becomes increasingly important. Millions of lives are lost annually due to local extreme heat and air pollution. We believe that the enhanced geospatial region representation and data efficiency provided by our approach will support more effective human health protection measures and inform geopolitical decision-making, leading to improved urban and environmental management as well as strategies for mitigating extreme climate effects.

\section{Conclusion and Future Work}
In this paper, we introduce \model, a space-aware method for socioeconomic indicator inference that employs a heterogeneous graph-based geospace representation framework. Our approach enhances indicator inference from environmental and societal perspectives, as well as higher-order cross-region correlations. Extensive experiments demonstrate that \model outperforms existing methods while maintaining strong performance even with limited data. Its minimal data reliance and double deployment strategies facilitate easy implementation for various domains. In the future, we plan to integrate advanced graph geometric techniques to accommodate more complex spatial structures within our graph framework, such as spherical space representation. This enhancement will allow the method to handle larger or global-scale inference where planar geometry is inadequate. Additionally, we aim to refine our code into a plug-and-play toolkit for indicator inference and interpolation, benefiting researchers in various fields.

\section{Acknowledgment}
This work is mainly supported by the Guangdong Basic and Applied Basic Research Foundation (No. 2025A1515011994). This work is also supported by the National Natural Science Foundation of China (No. 62402414), Guangzhou Municipal Science and Technology Project (No. 2023A03J0011), the Guangzhou Industrial Information and Intelligent Key Laboratory Project (No. 2024A03J0628), and a grant from State Key Laboratory of Resources and Environmental Information System, and Guangdong Provincial Key Lab of Integrated Communication, Sensing and Computation for Ubiquitous Internet of Things (No. 2023B1212010007). Additionally, this work benefits from the Red Bird MPhil Program at the Hong Kong University of Science and Technology (Guangzhou).
\bibliographystyle{ACM-Reference-Format}
\bibliography{sample-base}
\clearpage
\appendix
\begin{center}
    \LARGE{\textbf{{\textit{Appendix for:} ``Space-aware Socioeconomic Indicator Inference with Heterogeneous Graphs''}}}

\end{center}
\vspace{28pt}
\hrule

\section{More Introduction of ESA WorldCover Dataset}
\label{app:ESA}

\begin{figure}[h]
    \centering
    \includegraphics[width=0.9\linewidth]{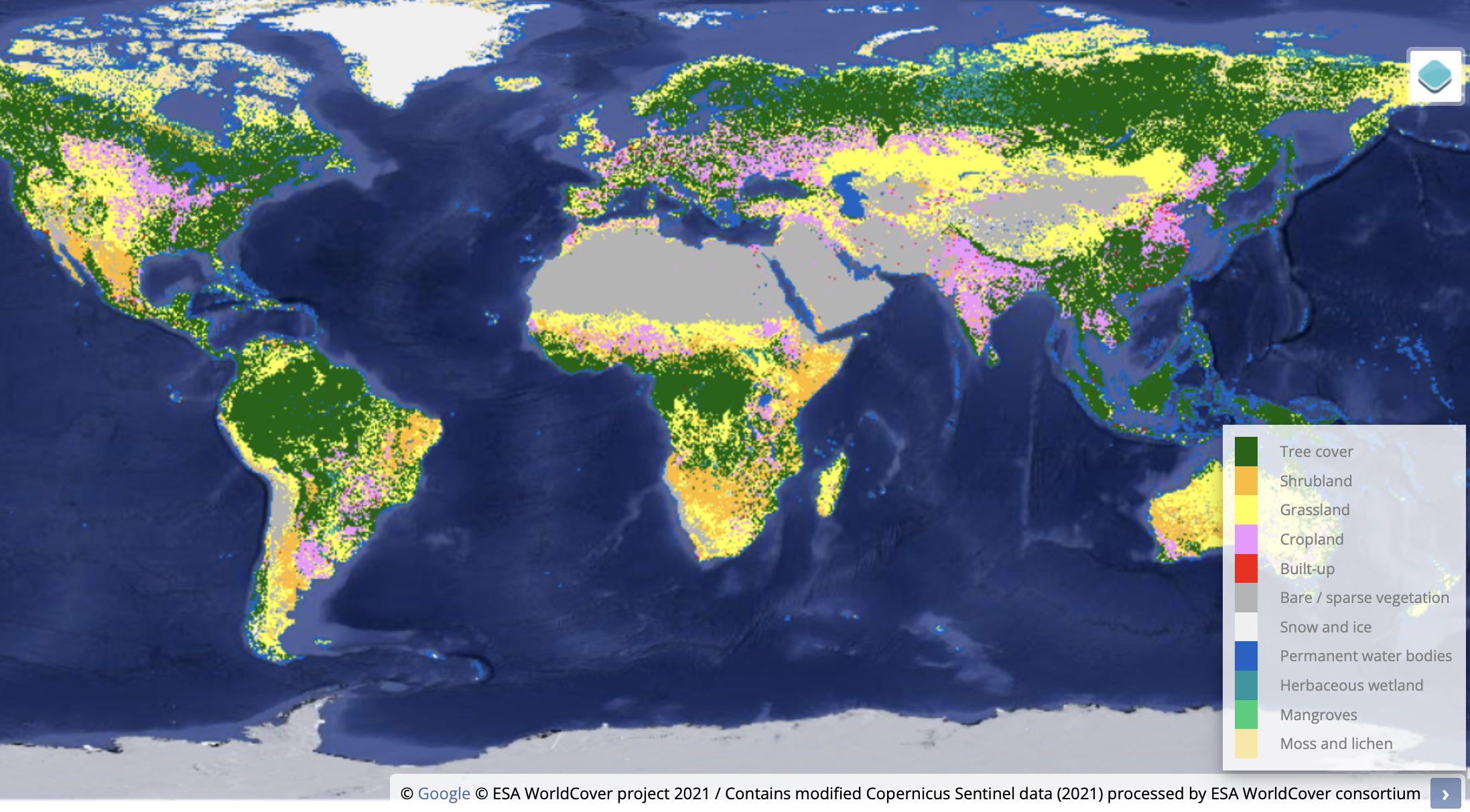}
    \vspace{-1em}
    \caption{Visualization of ESA WorldCover Dataset.}
    \label{fig:esa1}
\end{figure}

The ESA WorldCover dataset \footnote{https://esa-worldcover.org/en} represents a groundbreaking advancement in land use/land cover mapping, offering freely accessible, high-resolution (10 m) global coverage based on satellite imagery. Inspired by the 2017 WorldCover conference \footnote{https://worldcover2017.esa.int/}, the European Space Agency (ESA) launched the WorldCover project and the primary accomplishment of this endeavor was the introduction in October 2021 of a freely accessible global WorldCover Dataset at a groundbreaking 10 m resolution for the year 2020 \citep{zanaga2021esa,zanaga2022esa}. This dataset leverages satellite imagery from both Sentinel-1 and Sentinel-2, and encompasses 11 distinct geo-entity categories, shown in Figure \ref{fig:esa1}. It has also undergone rigorous independent validation by Wageningen University (for statistical accuracy) and the International Institute for Applied Systems Analysis (IIASA) (for spatial accuracy), attaining a notable global overall accuracy of approximately 75\% \citep{venter2022global}.

\begin{figure}[h]
    \centering
    \includegraphics[width=1.0\linewidth]{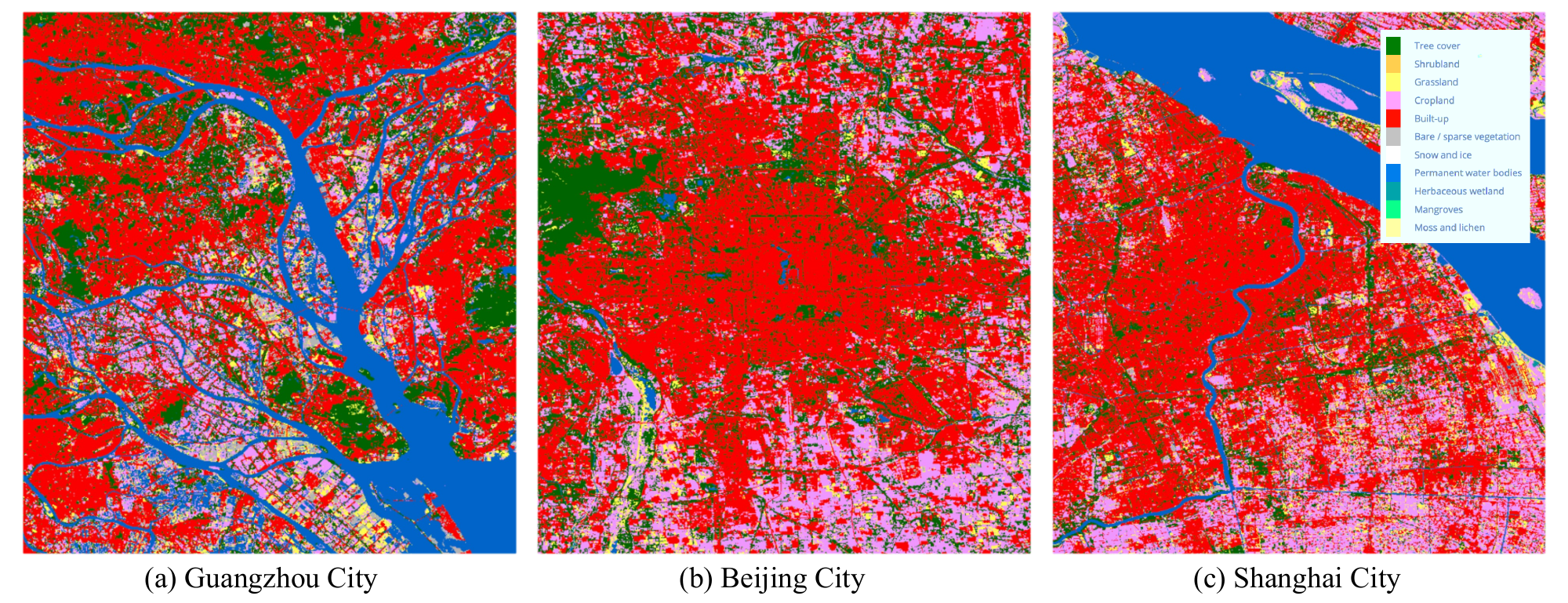}
    \vspace{-2em}
    \caption{Examples of Utilized Data in Our Paper.}
    \label{fig:esa2}
\end{figure}

WorldCover dataset is continuously updated and revised by ESA with a highly efficient processing pipeline and scalable infrastructure. The core model is trained on a total of 2,160,210 Sentinel-2 images and is able to process the whole world in less than 5 days \citep{zanaga2021esa,zanaga2022esa}. The revised version we utilized in this paper was released on 28 October 2022, which elevated the global overall accuracy to 76.7\% \citep{zanaga2022esa}. This version is free of charge to the entire community and widely accepted by the United Nations Convention to Combat Desertification (UNCCD), the World Resources Institute (WRI), the Centre for International Forestry Research (CIFOR), the Food and Agriculture Organization (FAO) and the Organisation for Economic Co-operation and Development (OECD). We visualized 3 examples of the WorldCover data unitized in this paper in Figure \ref{fig:esa2}.

\section{Evaluation metrics.} We employ Mean Absolute Error (MAE), rooted mean squared error (RMSE), and coefficient of determination ($R^2$) as evaluation metrics, these metrics are calculated as below:

\begin{equation}
\begin{aligned}
\operatorname{MAE}(y, \hat{y}) & =\frac{1}{|y|} \sum_{i=1}^{|y|}\left|y_{i}-\hat{y}_{i}\right|, \\
\end{aligned}    
\end{equation}

\begin{equation}
    \begin{aligned}
        \operatorname{RMSE}(y, \hat{y}) & =\sqrt{\frac{1}{|y|} \sum_{i=1}^{|y|}\left(y_{i}-\hat{y}_{i}\right)^{2}}
    \end{aligned}
\end{equation}
\begin{equation}
\begin{aligned}
    \operatorname{R^{2}} & = 1-\frac{\sum_{i=1}^{n}\left(y_{i}-\hat{y}_{i}\right)^{2}}{\sum_{i=1}^{n}\left(y_{i}-\bar{y}\right)^{2}}
    \end{aligned}
\end{equation}

\end{document}